\documentclass[10pt,twocolumn,letterpaper]{article}

\usepackage{iccv}
\usepackage{times}
\usepackage{epsfig}
\usepackage{graphicx}
\usepackage{amsmath}
\usepackage{amssymb}

\usepackage{booktabs}
\usepackage{subcaption}
\usepackage{scalerel}
\usepackage[export]{adjustbox}

\usepackage[pagebackref=true,breaklinks=true,letterpaper=true,colorlinks,bookmarks=false]{hyperref}

\iccvfinalcopy %

\ificcvfinal\fi

\begin{document}

\title{SparseGNV: Generating Novel Views of Indoor Scenes with Sparse Input Views}

\author{Weihao Cheng \quad Yan-Pei Cao \quad Ying Shan\\
ARC Lab, Tencent PCG\\
{\tt\small whcheng@tencent.com} \quad {\tt\small caoyanpei@gmail.com} \quad {\tt\small yingsshan@tencent.com}
}

\maketitle
\ificcvfinal\thispagestyle{empty}\fi

\begin{abstract}
We study to generate novel views of indoor scenes given sparse input views. The challenge is to achieve both photorealism and view consistency. We present SparseGNV: a learning framework that incorporates 3D structures and image generative models to generate novel views with three modules. The first module builds a neural point cloud as underlying geometry, providing contextual information and guidance for the target novel view. The second module utilizes a transformer-based network to map the scene context and the guidance into a shared latent space and autoregressively decodes the target view in the form of discrete image tokens. The third module reconstructs the tokens into the image of the target view. SparseGNV is trained across a large indoor scene dataset to learn generalizable priors. Once trained, it can efficiently generate novel views of an unseen indoor scene in a feed-forward manner. We evaluate SparseGNV on both real-world and synthetic indoor scenes and demonstrate that it outperforms state-of-the-art methods based on either neural radiance fields or conditional image generation. \url{https://github.com/xt4d/SparseGNV}
\end{abstract}

\section{Introduction}
\label{sec:intro}
Synthesizing high-quality novel views of 3D indoor scenes is a long-standing and challenging task in computer vision \cite{hedman2016scalable,philip2021free,gss}. Typically, this task requires dense scans from various viewpoints as input. However, indoor scenes are often spatially complex, and capturing every region of a scene can be expensive and even intractable. To overcome this challenge, we aim to synthesize novel views with sparse input observations, which reduces the data capture burden. An ideal approach should be capable of generating views by hallucinating unobserved regions with view consistency.

Sparse view synthesis methods have gained significant attention recently, particularly those based on neural radiance fields (NeRFs)~\cite{ddpnerf,regnerf}, which rely on a certain level of view coverage as input.
However, due to the lack of image generation ability, the above methods are intractable for hallucinating largely unobserved areas. Transformer-based methods~\cite{srt,viewformer} learn latent scene representations from 2D observations and conditionally generate images given new viewpoints. However, the lack of explicit 3D representation makes it challenging for these methods to synthesize visual details from unstructured latent space. Another line of work~\cite{synsin,pixelsynth,lookout} focuses on generating novel views or long-term videos starting from a single image, using generative networks to paint the ``outside'' of a view autoregressively, but they face limitations in synthesizing consistent views between multiple frames.
This leads to the core motivation of our approach: marrying explicit 3D scene structures with image generative models for a joint capability of generating views with a limited visual clue and maintaining scene consistency. 

\begin{figure}[t]
\begin{center}
\centering
\includegraphics[width=.99\linewidth]{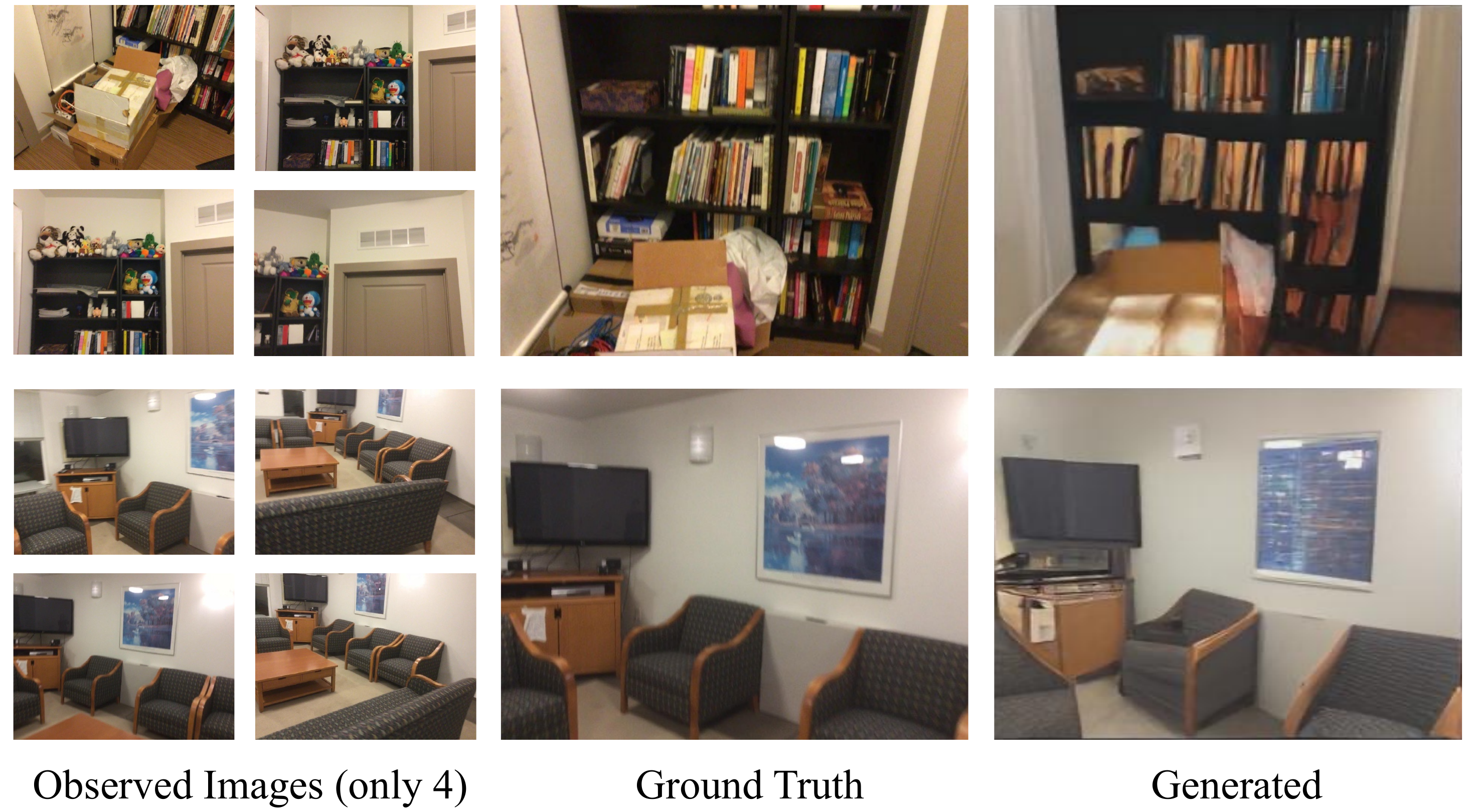}
\end{center}
\caption{
The proposed SparseGNV generates novel view images of unseen indoor scenes based on 4 observed views.
}
\label{fig:demo}
\end{figure}

We propose SparseGNV: a framework that learns generalizable scene priors to generate novel views conditioned on sparse input views. SparseGNV is first trained on a large indoor scene dataset to obtain priors that are generalizable across scenes. Once trained, SparseGNV can efficiently generate novel views in a forward pass given observed views of a new scene and target viewpoints, without the need for per-scene optimization. To generate 2D novel views grounded in 3D scene structures, we design SparseGNV with three modules: a neural geometry module, a view generator module, and an image converter module. 

\emph{The neural geometry module} reconstructs a set of input views into a 3D neural point cloud where each point is associated with an embedding vector. The neural point cloud can be rendered to 2D color and mask images from arbitrary viewpoints using volume rendering following Point-NeRF~\cite{pointnerf}. Although the point cloud can be scattered and incomplete due to input sparsity, the rendered images still provide structural and texture clues for hallucinating unobserved regions and maintaining consistency. 
\emph{The view generator module} generates a novel view conditioned on a \emph{scene context} and a \emph{query}. The \emph{scene context} is an overview of the given scene, which consists of the observed images and images rendered by the neural geometry module from multiple sampled viewpoints. It provides a global context that benefits inferring missing regions and maintaining consistency. The \emph{query} specifies the view that is required to generate. It consists of the rendered image from the target viewpoint. The \emph{query} provides guidance to retrieve information from the \emph{scene context} for generating the target novel view.
The module uses a joint convolution and transformer-based encoder-decoder network that maps the \emph{scene context} and the \emph{query} to a shared latent space, and then autoregressively generates the novel view in the form of discrete tokens~\cite{vqvae,vqgan}. 
\emph{The image converter module} is a convolutional decoder network that can reconstruct the discrete tokens back to photorealistic 2D images.

We evaluate SparseGNV on both real-world and synthetic indoor datasets, and the results outperform recent baselines using either neural radiance fields or conditional image generation. We show example generations of SparseGNV in Figure~\ref{fig:demo}.

\paragraph{Contributions}
\begin{itemize}
\item We propose SparseGNV: a learning framework to synthesize consistent novel views of indoor scenes with sparse input views, which combines neural 3D geometry and image generation model to enable photorealistic view synthesis with consistent structure faithful to the observations.
\item We design a joint convolution and transformer-based image generation network that effectively incorporates contextual information from 3D scene structures.
\item Evaluation results on real-world and synthetic indoor scene datasets demonstrate that SparseGNV achieves state-of-the-art performance for novel view synthesis with only a few observations.
\end{itemize}

\section{Related Work}
\label{sec:related}

\paragraph{Novel View Synthesis} 

Novel view synthesis is a task to produce images of scenes from arbitrary viewpoints given a number of input views. Early work achieves photorealistic synthesis by capturing a dense set of views \cite{MarcLevoy1996LightFR, StevenJGortler1996TheL}. Recently, neural networks based methods have made significant progress on enabling better synthesis quality, wider ranges of novel viewpoints, and more compact model representation. Neural radiance fields (NeRF) \cite{nerf} is a milestone work that trains multi-layer perceptron (MLP) to encode radiance and density for producing novel views via volume rendering. Following work based on NeRF extends novel view synthesis on varies of aspects: relaxing image constraints \cite{nerfw}, improving quality \cite{mipnerf}, dynamic view synthesis \cite{nsff, dnerf}, pose estimation \cite{barf,gnerf}, rendering in real-time \cite{plenoctrees}, and object / scene generation \cite{dreamfusion,dreamfields}. 

High-quality synthesis of scene views generally requires iterative per-scene optimizations with large number of observations. As dense inputs is unavailable in many scenarios, the study of few view synthesis is growing rapidly \cite{srn,dietnerf,regnerf,infonerf,geoaug}, and one direction is to learn priors across scenes and predicts novel views \cite{pixelnerf,mvsnerf,ibrnet,srt,viewformer}. PixelNeRF \cite{pixelnerf} is a learning framework that conditions NeRF on one or few input images to predict continuous scene representation. MVSNeRF \cite{mvsnerf} learns a generic deep neural network that combines plane-swept cost volumes with volume rendering for constructing radiance fields. IBRNet \cite{ibrnet} is a network of MLP and ray transformer that estimates radiance and volume density from multiple source views. Scene Representation Transformer \cite{srt} combines convolutional network and transformers to encodes input images into latent scene representations and decodes novel views. ViewFormer \cite{viewformer} is another transformer based approach with two stages, where images are encoded into tokens via a codebook network in the first stage, and the tokens of novel views are generated autoregressively conditioned on the inputs in the second stage. Depth prior can be useful for novel view synthesis \cite{ddpnerf} which completes a dense depth map first to guide optimization of NeRF. However, these methods can have poor performance with inputs of large sparsity.

\paragraph{Indoor Scene Synthesis from Sparse Views}
Synthesizing novel view of indoor scenes is a practical task naturally challenged by data sparsity. With incomplete RGB-D scans, SPSG \cite{spsg} generates high-quality colored reconstructions of 3D scenes in the form of TSDF. It uses a self-supervised approach to learn geometry and color inpainting with adversarial and perceptual supervisions on the 2D renderings of the reconstructions. CompNVS \cite{compnvs} is a framework to synthesis novel views from RGB-D scans with largely incomplete scene coverage. It first encodes scans into neural voxel grids, and then uses a geometry predictor with texture inpainter to complete the grids with embedding. A neural render decodes the grids into images and refined via adversarial training. These geometry based methods requires depth scan and strong 3D completion modeling which are hardly adapted to various scenes. PixelSynth \cite{pixelsynth} synthesizes novel view of a single image by outpainting unobserved areas projected via 3D reasoning. LookOutsideRoom \cite{lookout} synthesizes long-term video from a single scene image base on an autoregressive transformer modeling consecutive frames. These single image based methods are unable to maintain consistency between observations. Pathdreamer \cite{pathdreamer} targets on generating panorama images at novel positions given one or a few observations. It consists of a structure generator and an image generator. The structure generator projects observations into 3D semantic geometry. The image generator uses SPADE network \cite{spade} to generate photorealistic views from panorama semantic maps. Pathdreamer focuses on panorama images and requires semantic labeling of indoor scene which cannot be applied conventionally.

\newcommand{\pos}{\text{pos}}
\newcommand{\lss}{\mathcal{L}}
\newcommand{\encoder}{\mathtt{Encoder}}
\newcommand{\decoder}{\mathtt{Decoder}}
\newcommand{\sle}{\mathcal{S}_{< t}}
\newcommand{\softmax}{\mathtt{softmax}}
\newcommand{\sigmoid}{\mathtt{sigmoid}}
\newcommand{\ov}{\mathcal{O}}
\newcommand{\nv}{\mathcal{X}}

\begin{figure*}[t]
\begin{center}
\centering
\includegraphics[width=.99\linewidth]{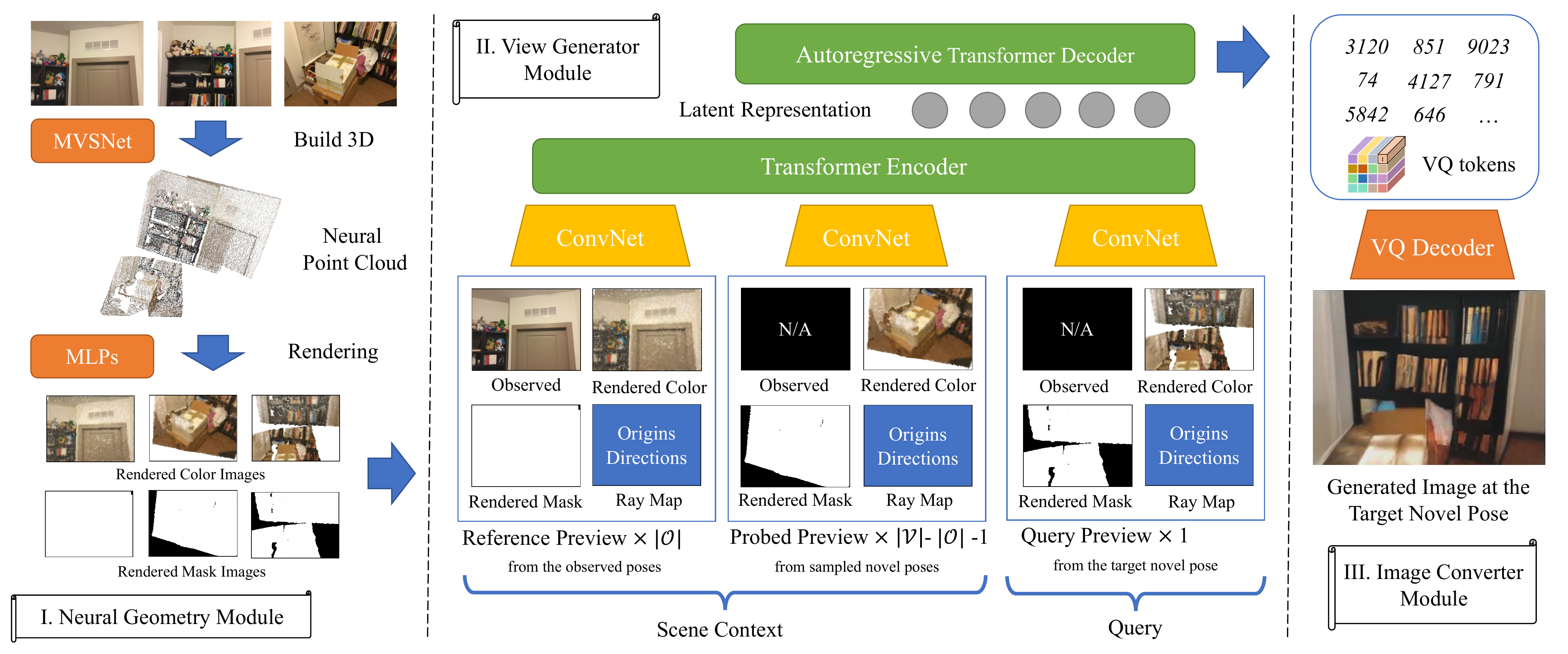}
\end{center}
\caption{
An overview of SparseGNV which consists of three modules: 1) Neural geometry module; 2) View generator Module; 3) Image Converter Module.
}
\label{fig:overview}
\end{figure*}

\section{Methodology}

In this section, we first briefly introduce the notation and the problem statement. We then propose SparseGNV with designs of the three modules. Lastly, we introduce the procedures of training and inference.

\subsection{Notation \& Problem}
Let $\mathcal{V}$ = $\{ (I_i, \pi_i) \,|\, i = 1, 2, ..., N\}$ be a set of views of indoor scenes, where $I_i \in \mathbb{R}^{W \times H \times 3}$ is the $i$-th color image and $\pi_i$ is the camera pose of $I_i$. $\mathcal{V}$ can be divided into an input observed view set $\ov$ and a novel view set $\nv$. Given an input sparse set of $\ov$, our problem is to generate a view image at a target novel viewpoint. As unobserved regions can be large, hallucinating novel views exactly matching ground truth is not easy. We therefore focus on high-fidelity generations while maintaining the view consistency.

\subsection{The SparseGNV framework}
We propose SparseGNV: a learning framework incorporating 3D scene structures and image generative models to generate consistent novel views of indoor scenes given only sparse input views.
SparseGNV is trained on a large indoor scene dataset to achieve generalization ability. 
Given sparse input views of an unseen scene, SparseGNV can efficiently generate novel views in a feed-forward manner. SparseGNV is designed with three modules: the neural geometry module, the view generator module, and the image converter module. The neural geometry module takes the input views to build a 3D neural point cloud~\cite{pointnerf} that can provide rendered guidance images from arbitrary viewpoints. The view generator module generates a novel view conditioned on a scene context of global information and a query regarding the target pose. The scene context and the query pack the information provided by the rendered guidance images, which are fed to a convolution and transformer-based network to generate the novel view in the form of discrete image tokens~\cite{vqvae,vqgan}. The image converter module reconstructs the tokens back to the final images through a decoder network. We show an overview of SparseGNV in Figure \ref{fig:overview}. The detailed description of the three modules is as follows. 
\\~
\textbf{Neural Geometry Module.} Given an input sparse set of observations $\ov$, the neural geometry module builds an underlying 3D neural point cloud that produces rendered guidance images from arbitrary poses. Those rendered guidance images provide structural and color clues that can complement scene representation and guide the generation of target novel views. 
\\~
The module builds a neural point cloud following Point-NeRF~\cite{pointnerf} with two steps: 1) reconstructs a 3D point cloud using the input $\ov$, which requires depths of the views that can be estimated via pre-trained Multi-View Stereo (MVS); 2) assigns each point of the cloud an embedding vector, which is computed by MVSNet \cite{mvsnet} given the corresponding pixel of the observed image. 
\\~
With the neural point cloud, the module can produce rendered color images $F_i \in \mathbb{R}^{W \times H \times 3}$ \cite{pointnerf}. In detail, given an arbitrary camera pose $\pi_i$, ray marching is performed to query a number of points on each ray. The embedding vectors of all the queried points are mapped to radiance and density via multi-layer perceptrons (MLPs). Through volume rendering, a ray color is obtained and assigned to the corresponding pixel of the image $F_i$. If a ray hits no neural point, the ray is marked as invalid. All the rays form a validation mask $M_i \in \{0, 1\}^{W \times H}$ indicating which part of $F_i$ is geometrically valid. The module output is formally expressed as:
\begin{equation}
	F_i, M_i = \mathtt{NeuralGeometry}(\pi_i, \ov \,;\, \theta),
\end{equation}
where $\theta$ is the parameters of module networks including the MVSNet and the MLPs. The mask $M_i$ can be used to filter out the invalid part of $F_i$ for a clear signal. 

The module networks are jointly trained to produce a visually reasonable $F_i$ with structure and color information. The objective is regressing $F_i$ to the ground truth color image $I_i$ on the valid rays:
\begin{equation}\label{eq:neural3d}
	\min_{\theta} \sum_{i} || ( F_i - I_i ) \odot M_i ||^2_2.
\end{equation}
\\~
\textbf{View Generator Module.} The view generator module uses a joint convolution and transformer-based network that takes a scene context and a query as input to generate a target novel view. The scene context is the global information that includes two types of \emph{``previews''}: reference previews and probed previews. The reference previews are from the input observed poses, and the probed previews are from several \emph{sampled} novel poses interpolated between the observed poses. The query is a preview from the target novel pose, which specifies the target viewpoint required to generate. Each preview of these three types is composed of four items: 1) an observed image $I_i$ (using an ``N/A'' image if unavailable); 2) a rendered color image $F_i$; 3) a mask image $M_i$; 4) a ray map $D_i$ of origins and directions derived from the camera pose $\pi_i$ \cite{srt}. For each preview, we concatenate the corresponding $I_i$, $F_i$, $M_i$, and $D_i$ to one multi-channel image, which is then fed into a convolutional network with the output spatially divided into a group of local patches $B_i$:
\begin{equation}
	B_i = \mathtt{ConvNet}(I_i \oplus F_i \oplus M_i \oplus D_i ).
\end{equation}
Each patch group $B_i$ is additionally labeled by adding a learnable segment embedding \cite{bert} regarding one of the three preview categories: reference, probed, and query. This allows the model to distinguish them and utilize information properly. We concatenate all the patches into one sequence, and pass it into a transformer encoder network to obtain a latent representation:
\begin{equation}
	h = \mathtt{TransformerEncoder}\left(\bigcup_i B_{i}\right).
\end{equation}
The latent representation $h$ is a set of hidden vectors that encodes both scene context and query information. The target novel view can then be generated conditioned on $h$. Due to the recent success of Vector Quantization (VQ) in image synthesis \cite{vqgan,dalle,lookout}, we present the target image as VQ codebook tokens $\mathcal{S}= \{s_1, s_2, ..., s_T \}$. The distribution of $\mathcal{S}$ is formulated as a probability $p(\mathcal{S} | h)$ which can be factorized as:
\begin{equation}\label{eq:factorize}
	p(\mathcal{S} | h) = \prod_{t=1}^{T} p(s_t | \sle, h),
\end{equation}
where $\sle$ = $\{s_1, s_2, ..., s_{t-1}\}$, and $p(s_t | \sle, h)$ is the probability of the $t$-th image token. We use a transformer decoder to model $p(\mathcal{S} | h)$ by autoregressively estimating $p(s_t | \sle, h)$. In detail, the last layer of the decoder generates hidden states $z$, and a linear layer $f(z)$ maps $z$ into a vector with the dimension of the codebook size. The probability $p(s_t | \sle,  h)$ is computed as $\softmax(f(z))$. We train the entire network by minimizing the objective of negative log-likelihood loss on the probability estimation:
\begin{equation}\label{eq:loss}
	\mathcal{L} = \sum_{s_t \in \mathcal{S}} -\log p(s_t | \sle, h).
\end{equation}
\\~
\textbf{Image Converter Module.} The image converter module is structurally based on a convolutional autoencoder network that encodes an image into discrete representation and decodes it back to the image. In SparseGNV, the image converter module plays two roles: 1) encoding a ground truth color image $I$ into VQ codebook tokens $\mathcal{S}$ for training the view generation module; 2) decoding a generated $\mathcal{S}$ back to the image at inference. The architecture of the converter network follows VQ-GAN \cite{vqgan}.

\subsection{Training \& Inference}
The training of SparseGNV requires two stages. In the first stage, the neural geometry module and the image converter module are trained separately. Given a scan of indoor scene views $\mathcal{V} = \{I_i, \pi_i\}$, we randomly sample a set of views as observations to build a neural point cloud, and iterate $I_i$ from $\mathcal{V}$ to supervise the rendered images for training the MVSNet and the MLPs jointly, as shown in Equation \eqref{eq:neural3d}. The image converter module is trained by reconstruction of a collection of wild images including indoor scene views. In the second stage, we use the neural geometry module to produce scene context and query, and supervise the view generator module to generate VQ tokens of novel views obtained by the image converter module, as shown in Equation \eqref{eq:loss}. The inference of SparseGNV is straightforward. Taking a number of observed views, we use the neural geometry module to build a neural point cloud. We then produce scene context and query with rendered images from the neural point cloud, and pass them to the view generator network. With the output latent representation $h$, we autoregressively draw out the VQ tokens using multinomial sampling. Lastly, we use the image converter model to reconstruct the VQ tokens into the final image.

\section{Experiments}

\newcommand{\garbagescale}{0.135}
\newcommand{\insidescale}{0.98}

\begin{table*}[t]
	\centering
		\begin{tabular}{@{}l|ccc|ccc|ccc@{}}
			\toprule
			& \multicolumn{3}{c|}{$|\mathcal{O}|=2$} & \multicolumn{3}{c|}{$|\mathcal{O}|=4$} &  \multicolumn{3}{c}{$|\mathcal{O}|=8$} \\
			\midrule
			Method & PSNR $\uparrow$ & SSIM $\uparrow$ & LPIPS $\downarrow$ & PSNR $\uparrow$ & SSIM $\uparrow$ & LPIPS $\downarrow$ & PSNR $\uparrow$ & SSIM $\uparrow$ & LPIPS $\downarrow$ \\ 
			\midrule
			Point-NeRF \cite{pointnerf} & 9.606 & 0.375 & 0.689 & 11.004 & 0.364 & 0.680 & 13.495 & 0.435 & 0.617 \\
			PixelSynth \cite{pixelsynth} & 11.503 & 0.412 & 0.750 & 12.261 & 0.443 & 0.716 & 12.880 & 0.459 & 0.684 \\
			IBRNet \cite{ibrnet} & 11.739 & 0.400 & 0.725 & 12.823 & 0.450 & 0.717 & 14.099 & 0.524 & 0.702 \\
			ViewFormer \cite{viewformer} & 14.365 & 0.541 & 0.674 & 14.927 & 0.549 & 0.649 & 15.420 & 0.553 & 0.633 \\
			DDP-NeRF \cite{ddpnerf} & 14.281 & 0.451 & 0.712 & 15.799 & 0.495 & 0.630 & 17.491 & 0.567 & 0.554 \\
			Ours w/o ``Geometry'' & 13.157 & 0.426 & 0.699 & 15.124 & 0.553 & 0. 617 & 16.010 & 0.537 & 0.582  \\
			Ours & \textbf{15.248} & \textbf{0.555} & \textbf{0.530} & \textbf{16.240} & \textbf{0.563} & \textbf{0.495} & \textbf{17.894} & \textbf{0.585} & \textbf{0.451} \\
			\bottomrule
		\end{tabular}
	\caption{
		Quantitative results on the ScanNet test scenes. 
	} 
	\label{tab:primary}
\end{table*}

\begin{figure*}[t]
 \centering 
 \begin{subfigure}{\garbagescale\textwidth}
    \centering
    \includegraphics[width=\insidescale\textwidth]{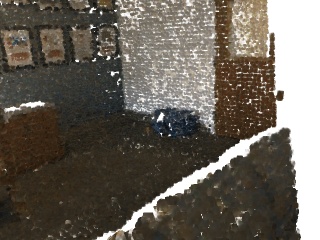}
     \captionsetup{labelformat=empty}
 \end{subfigure}
\begin{subfigure}{\garbagescale\textwidth}
	\centering
	\includegraphics[width=\insidescale\textwidth]{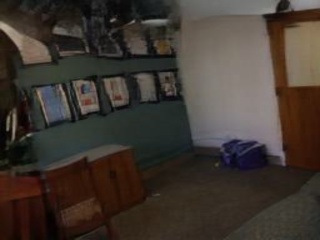}
	 \captionsetup{labelformat=empty}
\end{subfigure}
 \begin{subfigure}{\garbagescale\textwidth}
    \centering
    \includegraphics[width=\insidescale\textwidth]{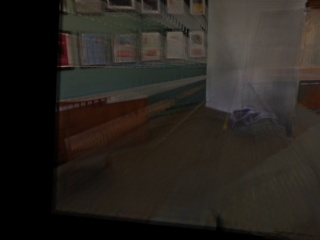}
     \captionsetup{labelformat=empty}
 \end{subfigure}
 \begin{subfigure}{\garbagescale\textwidth}
    \centering
    \includegraphics[width=\insidescale\textwidth]{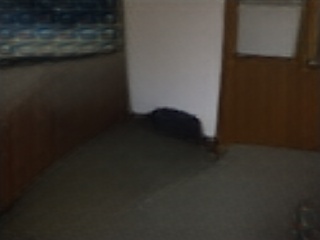}
     \captionsetup{labelformat=empty}
 \end{subfigure}
 \begin{subfigure}{\garbagescale\textwidth}
    \centering
    \includegraphics[width=\insidescale\textwidth]{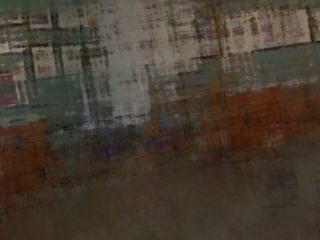}
    \captionsetup{labelformat=empty}
 \end{subfigure}
 \begin{subfigure}{\garbagescale\textwidth}
	\centering
	\includegraphics[width=\insidescale\textwidth]{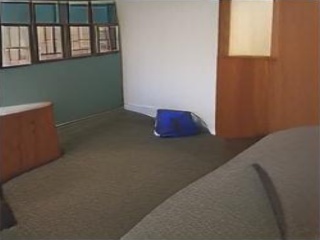}
	\captionsetup{labelformat=empty}
\end{subfigure}
 \begin{subfigure}{\garbagescale\textwidth}
	\centering
	\includegraphics[width=\insidescale\textwidth]{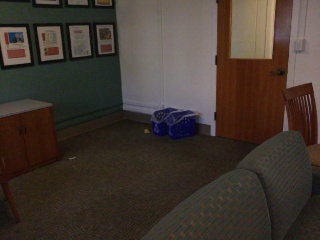}
	\captionsetup{labelformat=empty}
\end{subfigure}

 \vspace{4pt}

 \begin{subfigure}{\garbagescale\textwidth}
	\centering
	\includegraphics[width=\insidescale\textwidth]{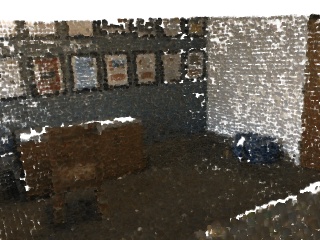}
	\captionsetup{labelformat=empty}
\end{subfigure}
\begin{subfigure}{\garbagescale\textwidth}
	\centering
	\includegraphics[width=\insidescale\textwidth]{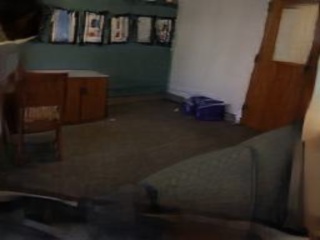}
	\captionsetup{labelformat=empty}
\end{subfigure}
\begin{subfigure}{\garbagescale\textwidth}
	\centering
	\includegraphics[width=\insidescale\textwidth]{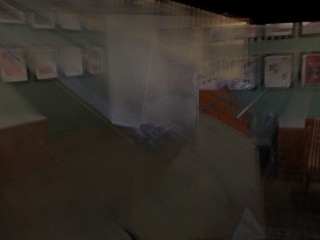}
	\captionsetup{labelformat=empty}
\end{subfigure}
\begin{subfigure}{\garbagescale\textwidth}
	\centering
	\includegraphics[width=\insidescale\textwidth]{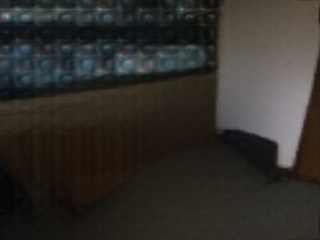}
	\captionsetup{labelformat=empty}
\end{subfigure}
\begin{subfigure}{\garbagescale\textwidth}
	\centering
	\includegraphics[width=\insidescale\textwidth]{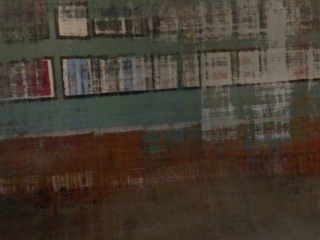}
	\captionsetup{labelformat=empty}
\end{subfigure}
\begin{subfigure}{\garbagescale\textwidth}
	\centering
	\includegraphics[width=\insidescale\textwidth]{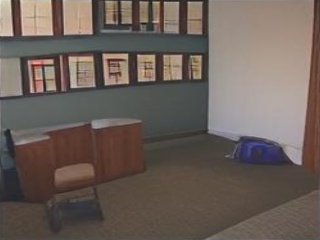}
	\captionsetup{labelformat=empty}
\end{subfigure}
\begin{subfigure}{\garbagescale\textwidth}
	\centering
	\includegraphics[width=\insidescale\textwidth]{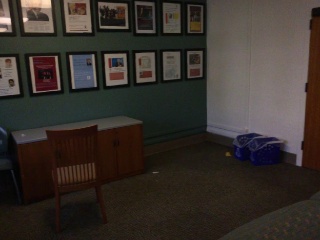}
	\captionsetup{labelformat=empty}
\end{subfigure}

 \vspace{6pt}

\begin{subfigure}{\garbagescale\textwidth}
	\centering
	\includegraphics[width=\insidescale\textwidth]{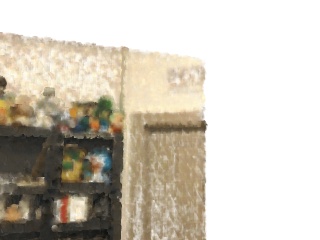}
	\captionsetup{labelformat=empty}
\end{subfigure}
\begin{subfigure}{\garbagescale\textwidth}
	\centering
	\includegraphics[width=\insidescale\textwidth]{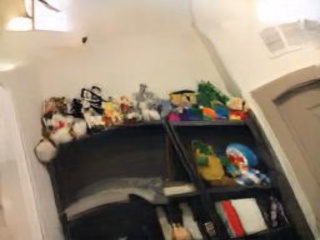}
	\captionsetup{labelformat=empty}
\end{subfigure}
\begin{subfigure}{\garbagescale\textwidth}
	\centering
	\includegraphics[width=\insidescale\textwidth]{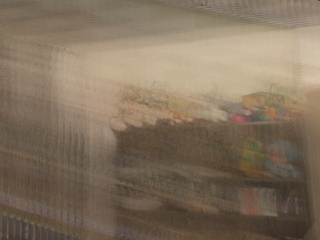}
	\captionsetup{labelformat=empty}
\end{subfigure}
\begin{subfigure}{\garbagescale\textwidth}
	\centering
	\includegraphics[width=\insidescale\textwidth]{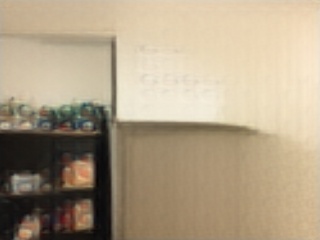}
	\captionsetup{labelformat=empty}
\end{subfigure}
\begin{subfigure}{\garbagescale\textwidth}
	\centering
	\includegraphics[width=\insidescale\textwidth]{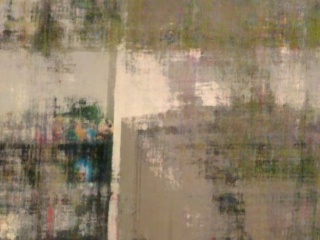}
	\captionsetup{labelformat=empty}
\end{subfigure}
\begin{subfigure}{\garbagescale\textwidth}
	\centering
	\includegraphics[width=\insidescale\textwidth]{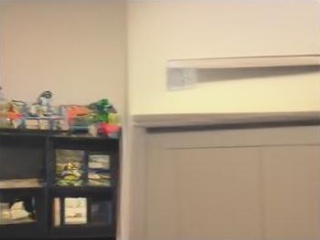}
	\captionsetup{labelformat=empty}
\end{subfigure}
\begin{subfigure}{\garbagescale\textwidth}
	\centering
	\includegraphics[width=\insidescale\textwidth]{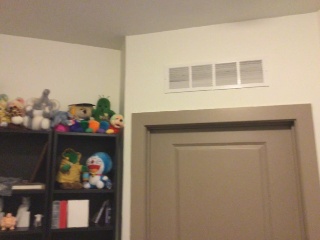}
	\captionsetup{labelformat=empty}
\end{subfigure}

 \vspace{4pt}

\begin{subfigure}{\garbagescale\textwidth}
	\centering
	\includegraphics[width=\insidescale\textwidth]{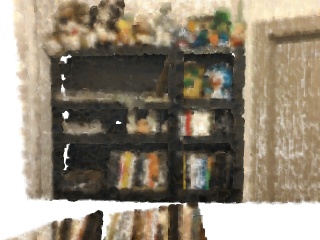}
	\captionsetup{labelformat=empty}
\end{subfigure}
\begin{subfigure}{\garbagescale\textwidth}
	\centering
	\includegraphics[width=\insidescale\textwidth]{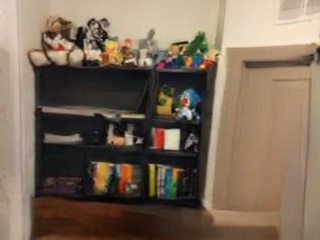}
	\captionsetup{labelformat=empty}
\end{subfigure}
\begin{subfigure}{\garbagescale\textwidth}
	\centering
	\includegraphics[width=\insidescale\textwidth]{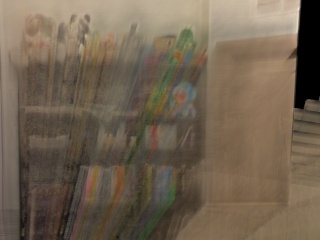}
	\captionsetup{labelformat=empty}
\end{subfigure}
\begin{subfigure}{\garbagescale\textwidth}
	\centering
	\includegraphics[width=\insidescale\textwidth]{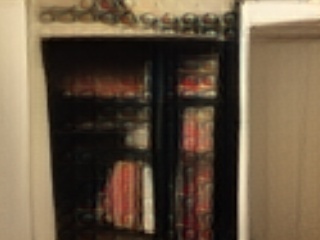}
	\captionsetup{labelformat=empty}
\end{subfigure}
\begin{subfigure}{\garbagescale\textwidth}
	\centering
	\includegraphics[width=\insidescale\textwidth]{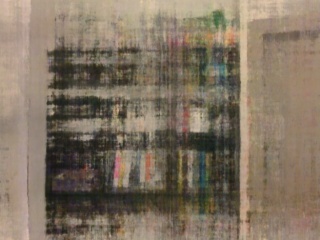}
	\captionsetup{labelformat=empty}
\end{subfigure}
\begin{subfigure}{\garbagescale\textwidth}
	\centering
	\includegraphics[width=\insidescale\textwidth]{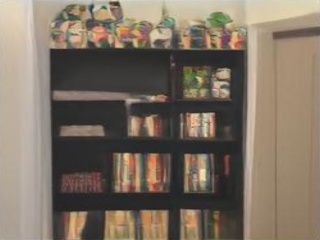}
	\captionsetup{labelformat=empty}
\end{subfigure}
\begin{subfigure}{\garbagescale\textwidth}
	\centering
	\includegraphics[width=\insidescale\textwidth]{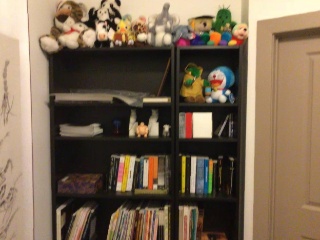}
	\captionsetup{labelformat=empty}
\end{subfigure}

\vspace{6pt}

\begin{subfigure}{\garbagescale\textwidth}
	\centering
	\includegraphics[width=\insidescale\textwidth]{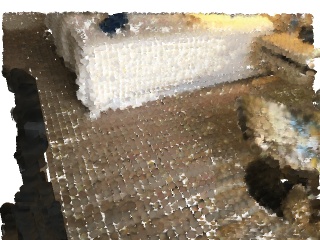}
	\captionsetup{labelformat=empty}
\end{subfigure}
\begin{subfigure}{\garbagescale\textwidth}
	\centering
	\includegraphics[width=\insidescale\textwidth]{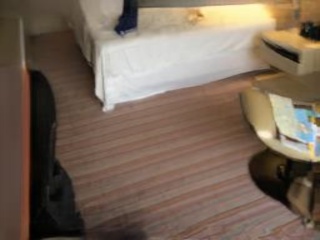}
	\captionsetup{labelformat=empty}
\end{subfigure}
\begin{subfigure}{\garbagescale\textwidth}
	\centering
	\includegraphics[width=\insidescale\textwidth]{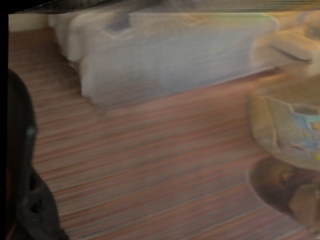}
	\captionsetup{labelformat=empty}
\end{subfigure}
\begin{subfigure}{\garbagescale\textwidth}
	\centering
	\includegraphics[width=\insidescale\textwidth]{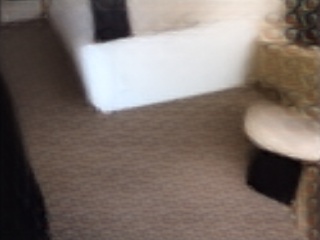}
	\captionsetup{labelformat=empty}
\end{subfigure}
\begin{subfigure}{\garbagescale\textwidth}
	\centering
	\includegraphics[width=\insidescale\textwidth]{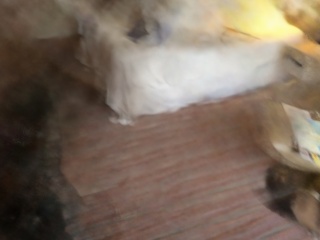}
	\captionsetup{labelformat=empty}
\end{subfigure}
\begin{subfigure}{\garbagescale\textwidth}
	\centering
	\includegraphics[width=\insidescale\textwidth]{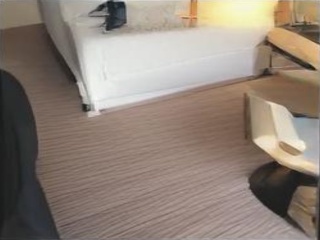}
	\captionsetup{labelformat=empty}
\end{subfigure}
\begin{subfigure}{\garbagescale\textwidth}
	\centering
	\includegraphics[width=\insidescale\textwidth]{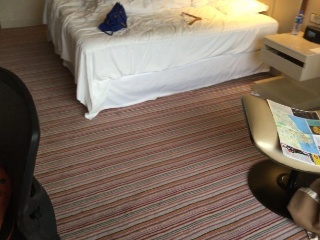}
	\captionsetup{labelformat=empty}
\end{subfigure}

\vspace{4pt}

\begin{subfigure}{\garbagescale\textwidth}
	\centering
	\includegraphics[width=\insidescale\textwidth]{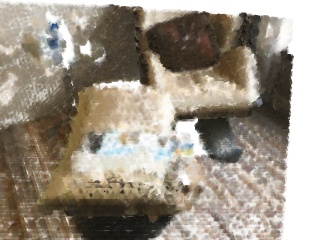}
	\captionsetup{labelformat=empty}
	\caption{Point-NeRF \cite{pointnerf}}
\end{subfigure}
\begin{subfigure}{\garbagescale\textwidth}
	\centering
	\includegraphics[width=\insidescale\textwidth]{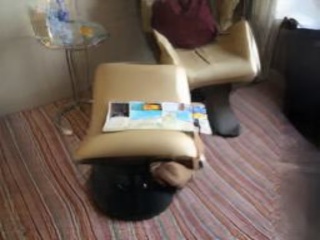}
	\captionsetup{labelformat=empty}
	\caption{PixelSynth \cite{pixelsynth}}
\end{subfigure}
\begin{subfigure}{\garbagescale\textwidth}
	\centering
	\includegraphics[width=\insidescale\textwidth]{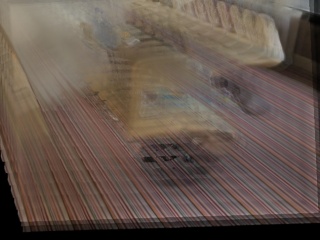}
	\captionsetup{labelformat=empty}
	\caption{IBRNet \cite{ibrnet}}
\end{subfigure}
\begin{subfigure}{\garbagescale\textwidth}
	\centering
	\includegraphics[width=\insidescale\textwidth]{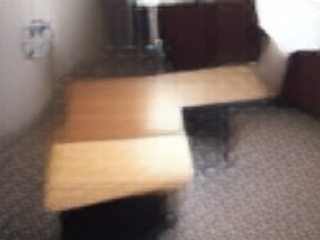}
	\captionsetup{labelformat=empty}
	\caption{ViewFormer \cite{viewformer}}
\end{subfigure}
\begin{subfigure}{\garbagescale\textwidth}
	\centering
	\includegraphics[width=\insidescale\textwidth]{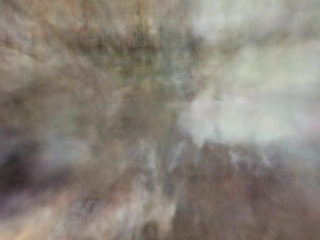}
	\captionsetup{labelformat=empty}
	\caption{DDP-NeRF \cite{ddpnerf}}
\end{subfigure}
\begin{subfigure}{\garbagescale\textwidth}
	\centering
	\includegraphics[width=\insidescale\textwidth]{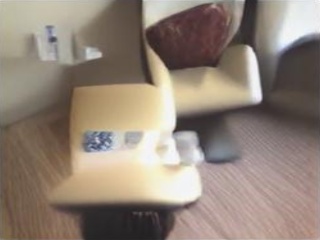}
	\captionsetup{labelformat=empty}
	\caption{Ours}
\end{subfigure}
\begin{subfigure}{\garbagescale\textwidth}
	\centering
	\includegraphics[width=\insidescale\textwidth]{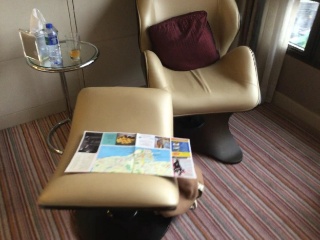}
	\captionsetup{labelformat=empty}
	\caption{Ground Truth}
\end{subfigure}

\vspace{-5pt}

 \caption{Synthesized novel views given input views of $|\mathcal{O}|$ = 2.} 

\vspace{-10pt}
 
 \label{fig:obs2}
\end{figure*}
\begin{figure*}[t]
 \centering 
 \begin{subfigure}{\garbagescale\textwidth}
    \centering
    \includegraphics[width=\insidescale\textwidth]{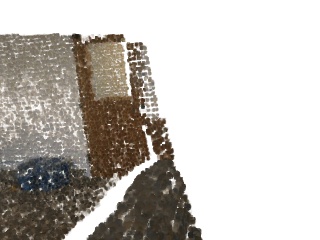}
     \captionsetup{labelformat=empty}
 \end{subfigure}
\begin{subfigure}{\garbagescale\textwidth}
	\centering
	\includegraphics[width=\insidescale\textwidth]{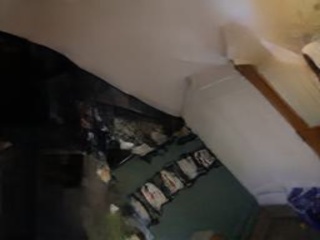}
	 \captionsetup{labelformat=empty}
\end{subfigure}
 \begin{subfigure}{\garbagescale\textwidth}
    \centering
    \includegraphics[width=\insidescale\textwidth]{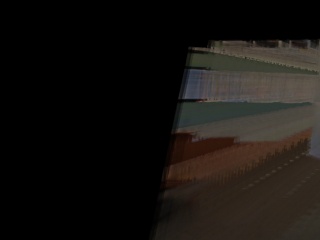}
     \captionsetup{labelformat=empty}
 \end{subfigure}
 \begin{subfigure}{\garbagescale\textwidth}
    \centering
    \includegraphics[width=\insidescale\textwidth]{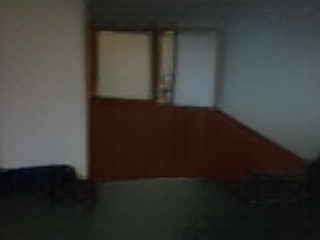}
     \captionsetup{labelformat=empty}
 \end{subfigure}
 \begin{subfigure}{\garbagescale\textwidth}
    \centering
    \includegraphics[width=\insidescale\textwidth]{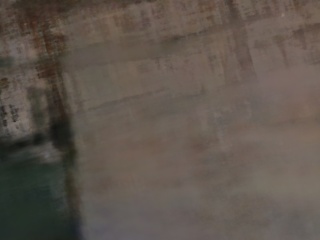}
    \captionsetup{labelformat=empty}
 \end{subfigure}
 \begin{subfigure}{\garbagescale\textwidth}
	\centering
	\includegraphics[width=\insidescale\textwidth]{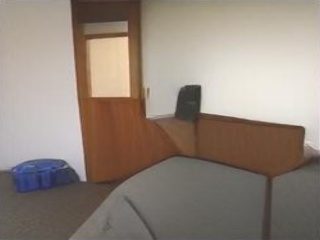}
	\captionsetup{labelformat=empty}
\end{subfigure}
 \begin{subfigure}{\garbagescale\textwidth}
	\centering
	\includegraphics[width=\insidescale\textwidth]{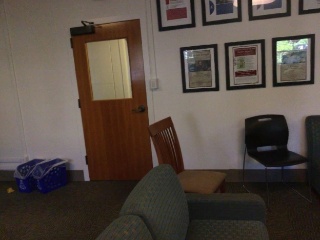}
	\captionsetup{labelformat=empty}
\end{subfigure}

 \vspace{4pt}

 \begin{subfigure}{\garbagescale\textwidth}
	\centering
	\includegraphics[width=\insidescale\textwidth]{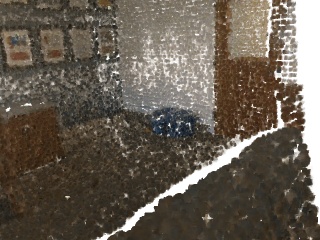}
	\captionsetup{labelformat=empty}
\end{subfigure}
\begin{subfigure}{\garbagescale\textwidth}
	\centering
	\includegraphics[width=\insidescale\textwidth]{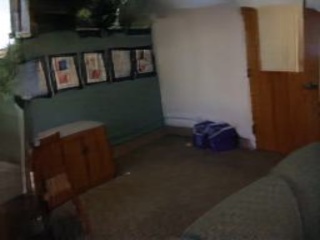}
	\captionsetup{labelformat=empty}
\end{subfigure}
\begin{subfigure}{\garbagescale\textwidth}
	\centering
	\includegraphics[width=\insidescale\textwidth]{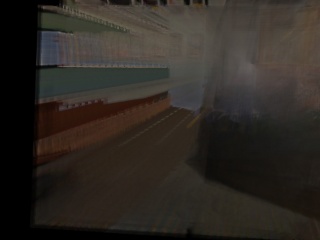}
	\captionsetup{labelformat=empty}
\end{subfigure}
\begin{subfigure}{\garbagescale\textwidth}
	\centering
	\includegraphics[width=\insidescale\textwidth]{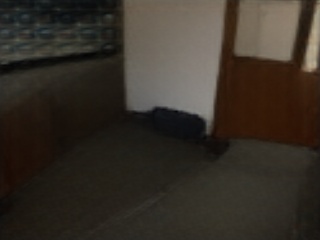}
	\captionsetup{labelformat=empty}
\end{subfigure}
\begin{subfigure}{\garbagescale\textwidth}
	\centering
	\includegraphics[width=\insidescale\textwidth]{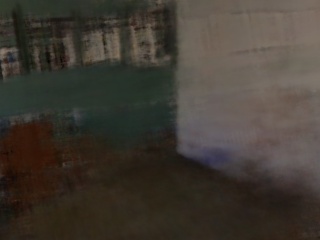}
	\captionsetup{labelformat=empty}
\end{subfigure}
\begin{subfigure}{\garbagescale\textwidth}
	\centering
	\includegraphics[width=\insidescale\textwidth]{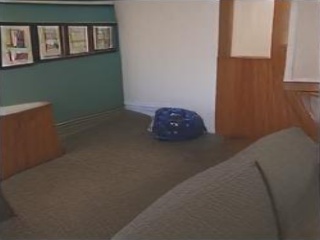}
	\captionsetup{labelformat=empty}
\end{subfigure}
\begin{subfigure}{\garbagescale\textwidth}
	\centering
	\includegraphics[width=\insidescale\textwidth]{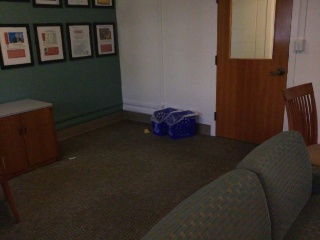}
	\captionsetup{labelformat=empty}
\end{subfigure}

 \vspace{6pt}

\begin{subfigure}{\garbagescale\textwidth}
	\centering
	\includegraphics[width=\insidescale\textwidth]{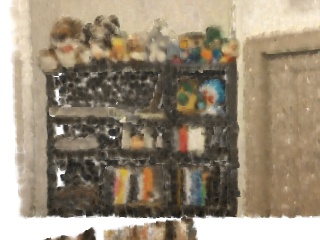}
	\captionsetup{labelformat=empty}
\end{subfigure}
\begin{subfigure}{\garbagescale\textwidth}
	\centering
	\includegraphics[width=\insidescale\textwidth]{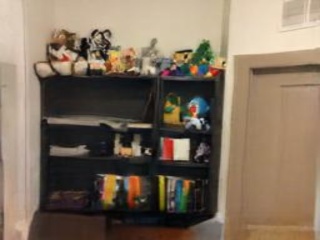}
	\captionsetup{labelformat=empty}
\end{subfigure}
\begin{subfigure}{\garbagescale\textwidth}
	\centering
	\includegraphics[width=\insidescale\textwidth]{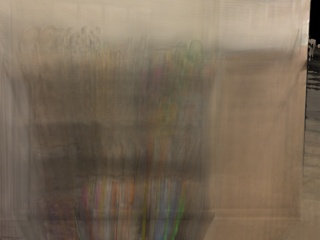}
	\captionsetup{labelformat=empty}
\end{subfigure}
\begin{subfigure}{\garbagescale\textwidth}
	\centering
	\includegraphics[width=\insidescale\textwidth]{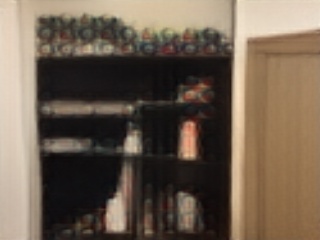}
	\captionsetup{labelformat=empty}
\end{subfigure}
\begin{subfigure}{\garbagescale\textwidth}
	\centering
	\includegraphics[width=\insidescale\textwidth]{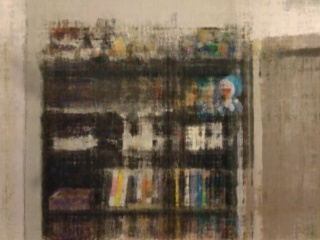}
	\captionsetup{labelformat=empty}
\end{subfigure}
\begin{subfigure}{\garbagescale\textwidth}
	\centering
	\includegraphics[width=\insidescale\textwidth]{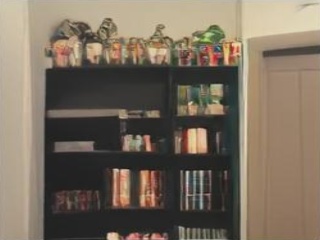}
	\captionsetup{labelformat=empty}
\end{subfigure}
\begin{subfigure}{\garbagescale\textwidth}
	\centering
	\includegraphics[width=\insidescale\textwidth]{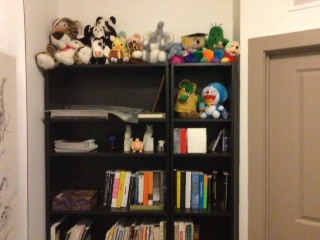}
	\captionsetup{labelformat=empty}
\end{subfigure}

 \vspace{4pt}

\begin{subfigure}{\garbagescale\textwidth}
	\centering
	\includegraphics[width=\insidescale\textwidth]{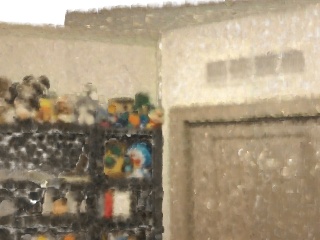}
	\captionsetup{labelformat=empty}
\end{subfigure}
\begin{subfigure}{\garbagescale\textwidth}
	\centering
	\includegraphics[width=\insidescale\textwidth]{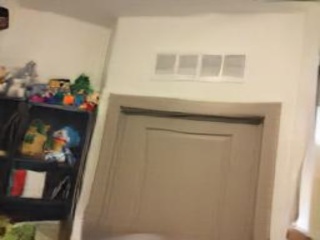}
	\captionsetup{labelformat=empty}
\end{subfigure}
\begin{subfigure}{\garbagescale\textwidth}
	\centering
	\includegraphics[width=\insidescale\textwidth]{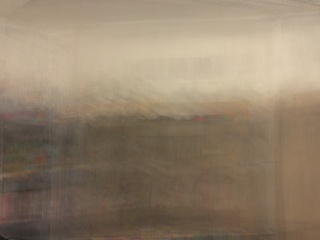}
	\captionsetup{labelformat=empty}
\end{subfigure}
\begin{subfigure}{\garbagescale\textwidth}
	\centering
	\includegraphics[width=\insidescale\textwidth]{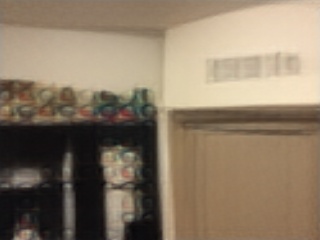}
	\captionsetup{labelformat=empty}
\end{subfigure}
\begin{subfigure}{\garbagescale\textwidth}
	\centering
	\includegraphics[width=\insidescale\textwidth]{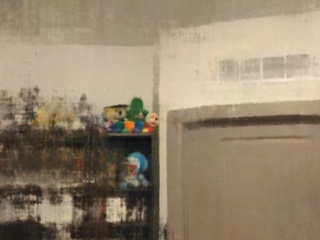}
	\captionsetup{labelformat=empty}
\end{subfigure}
\begin{subfigure}{\garbagescale\textwidth}
	\centering
	\includegraphics[width=\insidescale\textwidth]{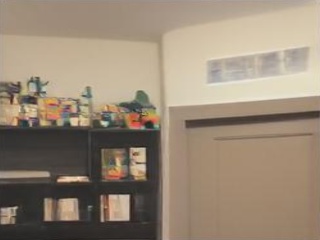}
	\captionsetup{labelformat=empty}
\end{subfigure}
\begin{subfigure}{\garbagescale\textwidth}
	\centering
	\includegraphics[width=\insidescale\textwidth]{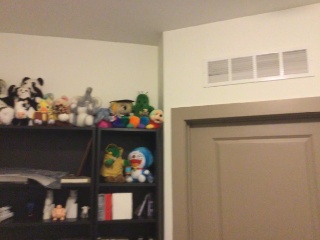}
	\captionsetup{labelformat=empty}
\end{subfigure}

\vspace{6pt}

\begin{subfigure}{\garbagescale\textwidth}
	\centering
	\includegraphics[width=\insidescale\textwidth]{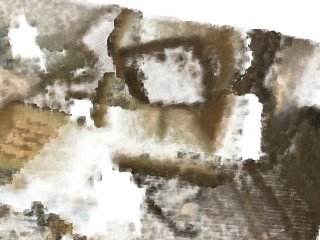}
	\captionsetup{labelformat=empty}
\end{subfigure}
\begin{subfigure}{\garbagescale\textwidth}
	\centering
	\includegraphics[width=\insidescale\textwidth]{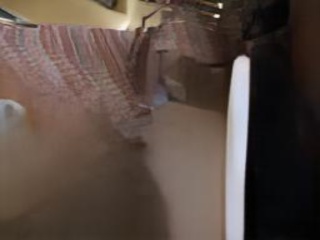}
	\captionsetup{labelformat=empty}
\end{subfigure}
\begin{subfigure}{\garbagescale\textwidth}
	\centering
	\includegraphics[width=\insidescale\textwidth]{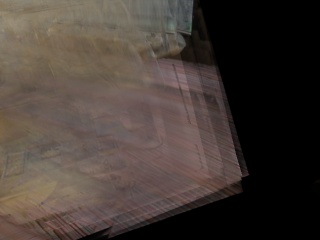}
	\captionsetup{labelformat=empty}
\end{subfigure}
\begin{subfigure}{\garbagescale\textwidth}
	\centering
	\includegraphics[width=\insidescale\textwidth]{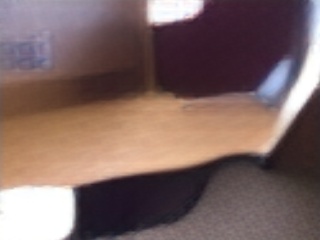}
	\captionsetup{labelformat=empty}
\end{subfigure}
\begin{subfigure}{\garbagescale\textwidth}
	\centering
	\includegraphics[width=\insidescale\textwidth]{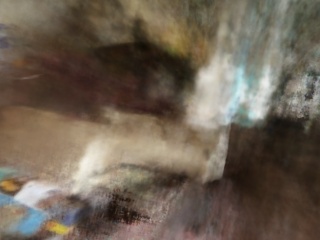}
	\captionsetup{labelformat=empty}
\end{subfigure}
\begin{subfigure}{\garbagescale\textwidth}
	\centering
	\includegraphics[width=\insidescale\textwidth]{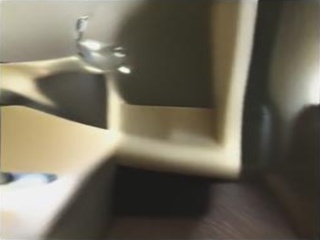}
	\captionsetup{labelformat=empty}
\end{subfigure}
\begin{subfigure}{\garbagescale\textwidth}
	\centering
	\includegraphics[width=\insidescale\textwidth]{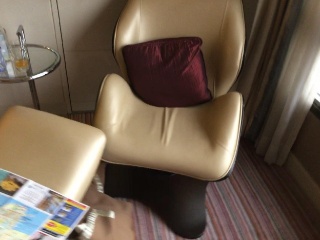}
	\captionsetup{labelformat=empty}
\end{subfigure}

\vspace{4pt}

\begin{subfigure}{\garbagescale\textwidth}
	\centering
	\includegraphics[width=\insidescale\textwidth]{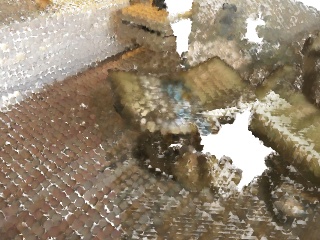}
	\captionsetup{labelformat=empty}
	\caption{Point-NeRF \cite{pointnerf}}
\end{subfigure}
\begin{subfigure}{\garbagescale\textwidth}
	\centering
	\includegraphics[width=\insidescale\textwidth]{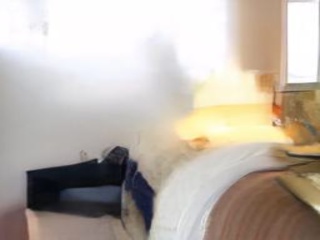}
	\captionsetup{labelformat=empty}
	\caption{PixelSynth \cite{pixelsynth}}
\end{subfigure}
\begin{subfigure}{\garbagescale\textwidth}
	\centering
	\includegraphics[width=\insidescale\textwidth]{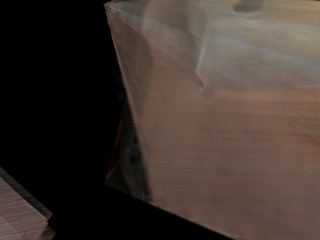}
	\captionsetup{labelformat=empty}
	\caption{IBRNet \cite{ibrnet}}
\end{subfigure}
\begin{subfigure}{\garbagescale\textwidth}
	\centering
	\includegraphics[width=\insidescale\textwidth]{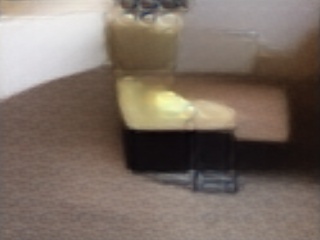}
	\captionsetup{labelformat=empty}
	\caption{ViewFormer \cite{viewformer}}
\end{subfigure}
\begin{subfigure}{\garbagescale\textwidth}
	\centering
	\includegraphics[width=\insidescale\textwidth]{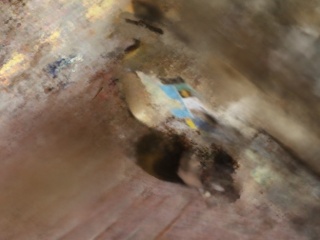}
	\captionsetup{labelformat=empty}
	\caption{DDP-NeRF \cite{ddpnerf}}
\end{subfigure}
\begin{subfigure}{\garbagescale\textwidth}
	\centering
	\includegraphics[width=\insidescale\textwidth]{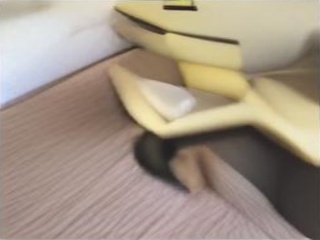}
	\captionsetup{labelformat=empty}
	\caption{Ours}
\end{subfigure}
\begin{subfigure}{\garbagescale\textwidth}
	\centering
	\includegraphics[width=\insidescale\textwidth]{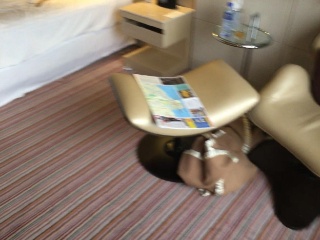}
	\captionsetup{labelformat=empty}
	\caption{Ground Truth}
\end{subfigure}

\vspace{-5pt}

 \caption{Synthesized novel views given input views of $|\mathcal{O}|$ = 4.} 

\vspace{-10pt}
 
 \label{fig:obs4}
\end{figure*}
\begin{figure*}[th]

\centering 
\begin{subfigure}{\garbagescale\textwidth}
    \centering
    \includegraphics[width=\insidescale\textwidth]{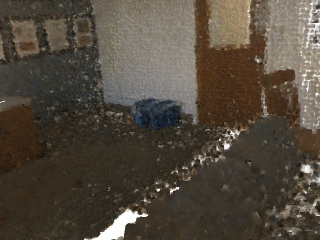}
    \captionsetup{labelformat=empty}
\end{subfigure}
\begin{subfigure}{\garbagescale\textwidth}
    \centering
    \includegraphics[width=\insidescale\textwidth]{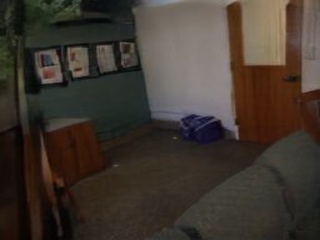}
    \captionsetup{labelformat=empty}
\end{subfigure}
\begin{subfigure}{\garbagescale\textwidth}
    \centering
    \includegraphics[width=\insidescale\textwidth]{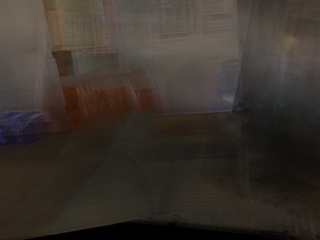}
    \captionsetup{labelformat=empty}
\end{subfigure}
\begin{subfigure}{\garbagescale\textwidth}
    \centering
    \includegraphics[width=\insidescale\textwidth]{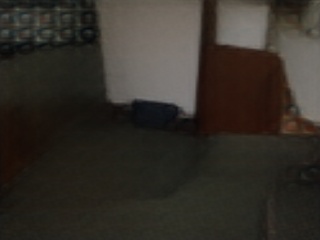}
    \captionsetup{labelformat=empty}
\end{subfigure}
\begin{subfigure}{\garbagescale\textwidth}
    \centering
    \includegraphics[width=\insidescale\textwidth]{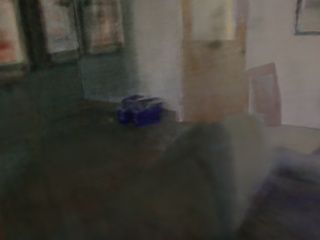}
    \captionsetup{labelformat=empty}
\end{subfigure}
\begin{subfigure}{\garbagescale\textwidth}
    \centering
    \includegraphics[width=\insidescale\textwidth]{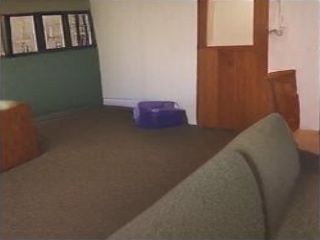}
    \captionsetup{labelformat=empty}
\end{subfigure}
\begin{subfigure}{\garbagescale\textwidth}
    \centering
    \includegraphics[width=\insidescale\textwidth]{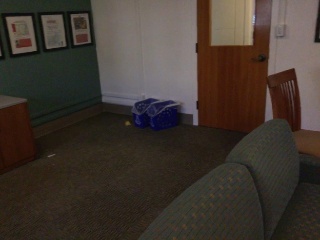}
    \captionsetup{labelformat=empty}
\end{subfigure}

\vspace{4pt}

\begin{subfigure}{\garbagescale\textwidth}
	\centering
	\includegraphics[width=\insidescale\textwidth]{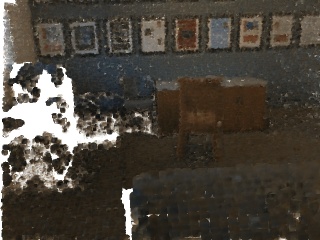}
	\captionsetup{labelformat=empty}
\end{subfigure}
\begin{subfigure}{\garbagescale\textwidth}
	\centering
	\includegraphics[width=\insidescale\textwidth]{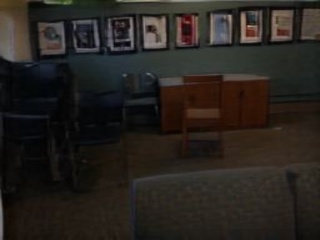}
	\captionsetup{labelformat=empty}
\end{subfigure}
\begin{subfigure}{\garbagescale\textwidth}
	\centering
	\includegraphics[width=\insidescale\textwidth]{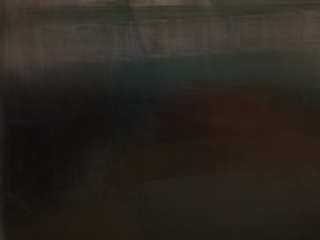}
	\captionsetup{labelformat=empty}
\end{subfigure}
\begin{subfigure}{\garbagescale\textwidth}
	\centering
	\includegraphics[width=\insidescale\textwidth]{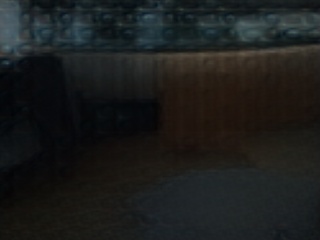}
	\captionsetup{labelformat=empty}
\end{subfigure}
\begin{subfigure}{\garbagescale\textwidth}
	\centering
	\includegraphics[width=\insidescale\textwidth]{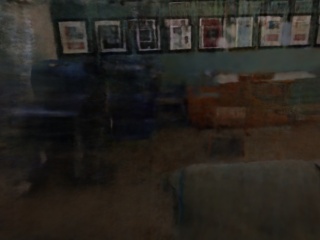}
	\captionsetup{labelformat=empty}
\end{subfigure}
\begin{subfigure}{\garbagescale\textwidth}
	\centering
	\includegraphics[width=\insidescale\textwidth]{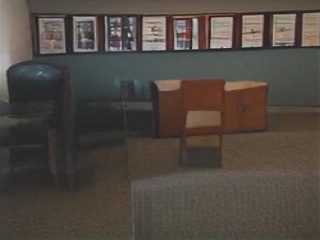}
	\captionsetup{labelformat=empty}
\end{subfigure}
\begin{subfigure}{\garbagescale\textwidth}
	\centering
	\includegraphics[width=\insidescale\textwidth]{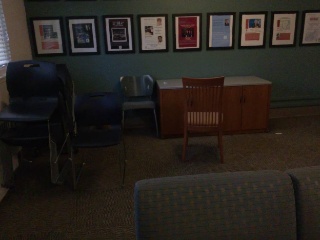}
	\captionsetup{labelformat=empty}
\end{subfigure}

 \vspace{6pt}

\begin{subfigure}{\garbagescale\textwidth}
	\centering
	\includegraphics[width=\insidescale\textwidth]{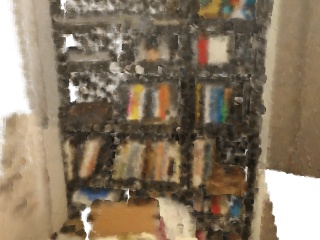}
	\captionsetup{labelformat=empty}
\end{subfigure}
\begin{subfigure}{\garbagescale\textwidth}
	\centering
	\includegraphics[width=\insidescale\textwidth]{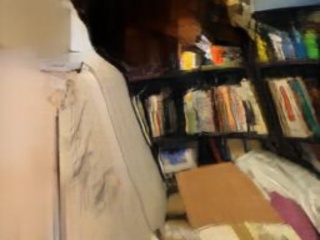}
	\captionsetup{labelformat=empty}
\end{subfigure}
\begin{subfigure}{\garbagescale\textwidth}
	\centering
	\includegraphics[width=\insidescale\textwidth]{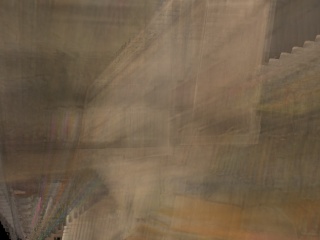}
	\captionsetup{labelformat=empty}
\end{subfigure}
\begin{subfigure}{\garbagescale\textwidth}
	\centering
	\includegraphics[width=\insidescale\textwidth]{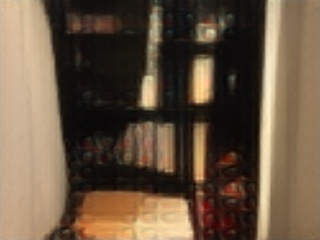}
	\captionsetup{labelformat=empty}
\end{subfigure}
\begin{subfigure}{\garbagescale\textwidth}
	\centering
	\includegraphics[width=\insidescale\textwidth]{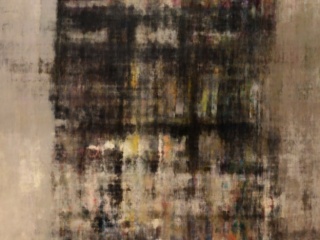}
	\captionsetup{labelformat=empty}
\end{subfigure}
\begin{subfigure}{\garbagescale\textwidth}
	\centering
	\includegraphics[width=\insidescale\textwidth]{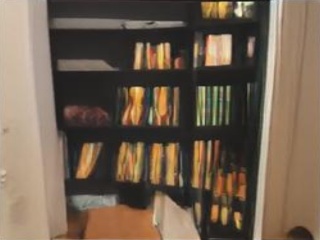}
	\captionsetup{labelformat=empty}
\end{subfigure}
\begin{subfigure}{\garbagescale\textwidth}
	\centering
	\includegraphics[width=\insidescale\textwidth]{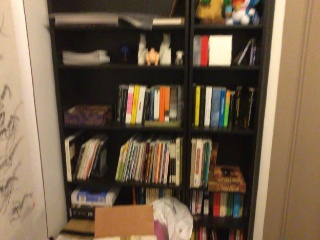}
	\captionsetup{labelformat=empty}
\end{subfigure}

 \vspace{4pt}

\begin{subfigure}{\garbagescale\textwidth}
	\centering
	\includegraphics[width=\insidescale\textwidth]{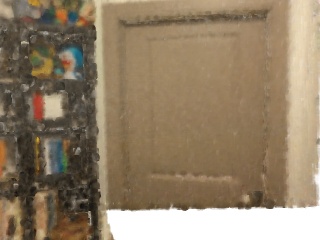}
	\captionsetup{labelformat=empty}
\end{subfigure}
\begin{subfigure}{\garbagescale\textwidth}
	\centering
	\includegraphics[width=\insidescale\textwidth]{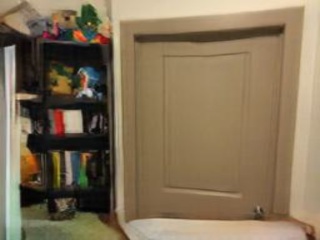}
	\captionsetup{labelformat=empty}
\end{subfigure}
\begin{subfigure}{\garbagescale\textwidth}
	\centering
	\includegraphics[width=\insidescale\textwidth]{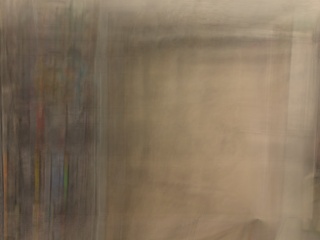}
	\captionsetup{labelformat=empty}
\end{subfigure}
\begin{subfigure}{\garbagescale\textwidth}
	\centering
	\includegraphics[width=\insidescale\textwidth]{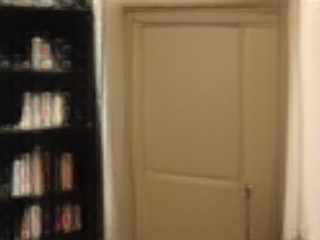}
	\captionsetup{labelformat=empty}
\end{subfigure}
\begin{subfigure}{\garbagescale\textwidth}
	\centering
	\includegraphics[width=\insidescale\textwidth]{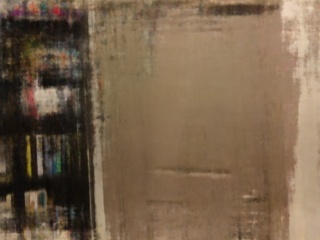}
	\captionsetup{labelformat=empty}
\end{subfigure}
\begin{subfigure}{\garbagescale\textwidth}
	\centering
	\includegraphics[width=\insidescale\textwidth]{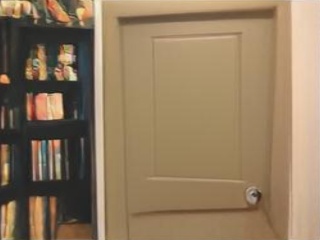}
	\captionsetup{labelformat=empty}
\end{subfigure}
\begin{subfigure}{\garbagescale\textwidth}
	\centering
	\includegraphics[width=\insidescale\textwidth]{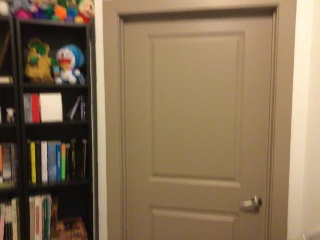}
	\captionsetup{labelformat=empty}
\end{subfigure}

\vspace{6pt}

\begin{subfigure}{\garbagescale\textwidth}
	\centering
	\includegraphics[width=\insidescale\textwidth]{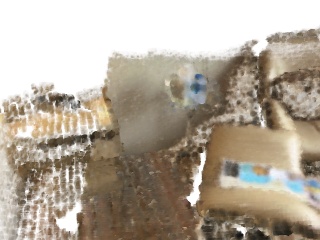}
	\captionsetup{labelformat=empty}
\end{subfigure}
\begin{subfigure}{\garbagescale\textwidth}
	\centering
	\includegraphics[width=\insidescale\textwidth]{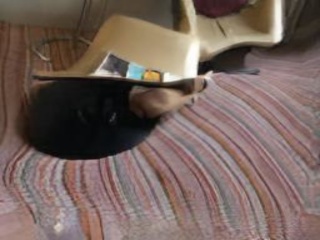}
	\captionsetup{labelformat=empty}
\end{subfigure}
\begin{subfigure}{\garbagescale\textwidth}
	\centering
	\includegraphics[width=\insidescale\textwidth]{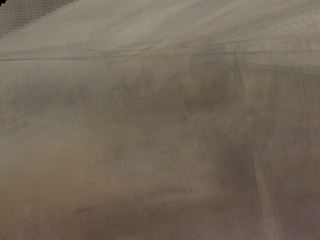}
	\captionsetup{labelformat=empty}
\end{subfigure}
\begin{subfigure}{\garbagescale\textwidth}
	\centering
	\includegraphics[width=\insidescale\textwidth]{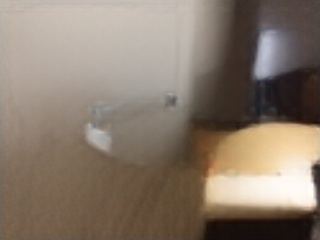}
	\captionsetup{labelformat=empty}
\end{subfigure}
\begin{subfigure}{\garbagescale\textwidth}
	\centering
	\includegraphics[width=\insidescale\textwidth]{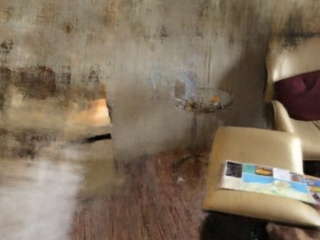}
	\captionsetup{labelformat=empty}
\end{subfigure}
\begin{subfigure}{\garbagescale\textwidth}
	\centering
	\includegraphics[width=\insidescale\textwidth]{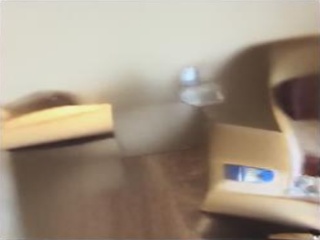}
	\captionsetup{labelformat=empty}
\end{subfigure}
\begin{subfigure}{\garbagescale\textwidth}
	\centering
	\includegraphics[width=\insidescale\textwidth]{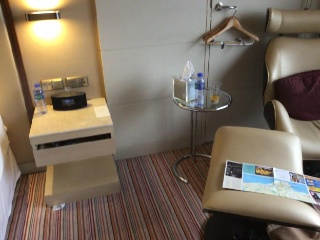}
	\captionsetup{labelformat=empty}
\end{subfigure}

\vspace{4pt}

\begin{subfigure}{\garbagescale\textwidth}
	\centering
	\includegraphics[width=\insidescale\textwidth]{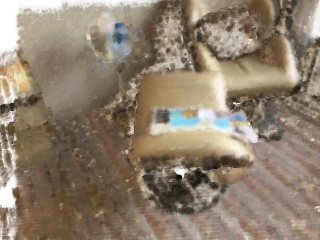}
	\captionsetup{labelformat=empty}
	\caption{Point-NeRF \cite{pointnerf}}
\end{subfigure}
\begin{subfigure}{\garbagescale\textwidth}
	\centering
	\includegraphics[width=\insidescale\textwidth]{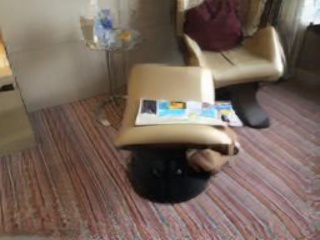}
	\captionsetup{labelformat=empty}
	\caption{PixelSynth \cite{pixelsynth}}
\end{subfigure}
\begin{subfigure}{\garbagescale\textwidth}
	\centering
	\includegraphics[width=\insidescale\textwidth]{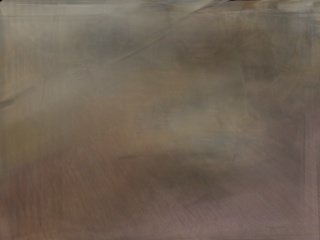}
	\captionsetup{labelformat=empty}
	\caption{IBRNet \cite{ibrnet}}
\end{subfigure}
\begin{subfigure}{\garbagescale\textwidth}
	\centering
	\includegraphics[width=\insidescale\textwidth]{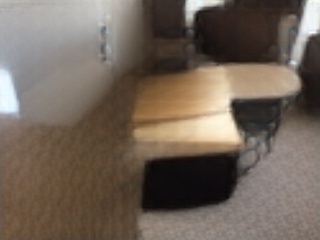}
	\captionsetup{labelformat=empty}
	\caption{ViewFormer \cite{viewformer}}
\end{subfigure}
\begin{subfigure}{\garbagescale\textwidth}
	\centering
	\includegraphics[width=\insidescale\textwidth]{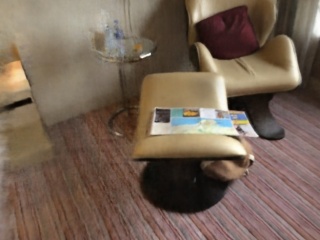}
	\captionsetup{labelformat=empty}
	\caption{DDP-NeRF \cite{ddpnerf}}
\end{subfigure}
\begin{subfigure}{\garbagescale\textwidth}
	\centering
	\includegraphics[width=\insidescale\textwidth]{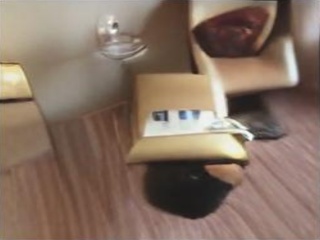}
	\captionsetup{labelformat=empty}
	\caption{Ours}
\end{subfigure}
\begin{subfigure}{\garbagescale\textwidth}
	\centering
	\includegraphics[width=\insidescale\textwidth]{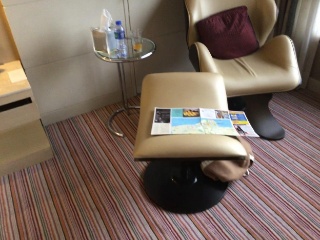}
	\captionsetup{labelformat=empty}
	\caption{Ground Truth}
\end{subfigure}

\vspace{-5pt}

 \caption{Synthesized novel views given input views of $|\mathcal{O}|$ = 8.} 
 \label{fig:obs8}
\end{figure*}
\begin{figure*}[th]
 \centering 
 \begin{subfigure}{\garbagescale\textwidth}
    \centering
    \includegraphics[width=\insidescale\textwidth,cfbox=red 1pt 0pt]{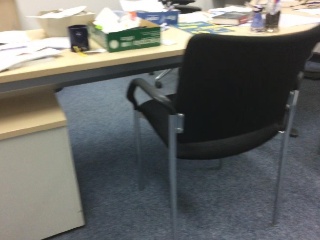}
     \captionsetup{labelformat=empty}
 \end{subfigure}
\begin{subfigure}{\garbagescale\textwidth}
	\centering
	\includegraphics[width=\insidescale\textwidth]{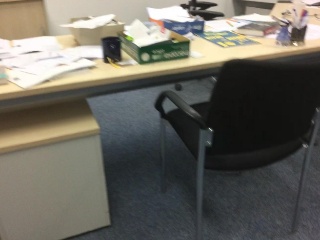}
	 \captionsetup{labelformat=empty}
\end{subfigure}
 \begin{subfigure}{\garbagescale\textwidth}
    \centering
    \includegraphics[width=\insidescale\textwidth]{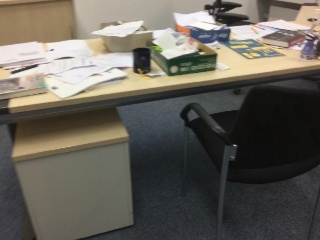}
     \captionsetup{labelformat=empty}
 \end{subfigure}
 \begin{subfigure}{\garbagescale\textwidth}
    \centering
    \includegraphics[width=\insidescale\textwidth]{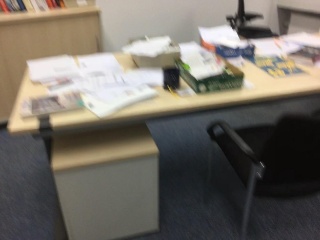}
     \captionsetup{labelformat=empty}
 \end{subfigure}
 \begin{subfigure}{\garbagescale\textwidth}
    \centering
    \includegraphics[width=\insidescale\textwidth]{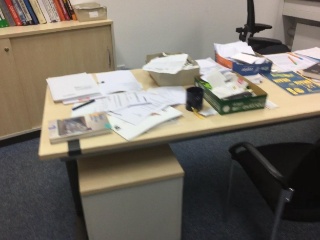}
    \captionsetup{labelformat=empty}
 \end{subfigure}
 \begin{subfigure}{\garbagescale\textwidth}
	\centering
	\includegraphics[width=\insidescale\textwidth]{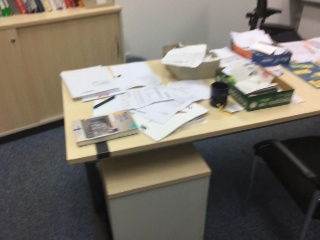}
	\captionsetup{labelformat=empty}
\end{subfigure}
 \begin{subfigure}{\garbagescale\textwidth}
	\centering
	\includegraphics[width=\insidescale\textwidth]{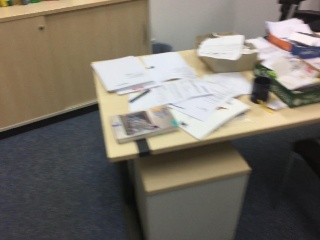}
	\captionsetup{labelformat=empty}
\end{subfigure}

 \vspace{3pt}

 \begin{subfigure}{\garbagescale\textwidth}
	\centering
	\includegraphics[width=\insidescale\textwidth]{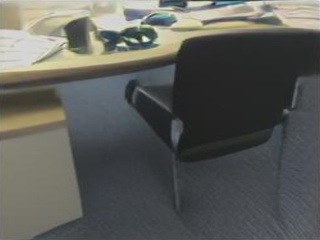}
	\captionsetup{labelformat=empty}
\end{subfigure}
\begin{subfigure}{\garbagescale\textwidth}
	\centering
	\includegraphics[width=\insidescale\textwidth]{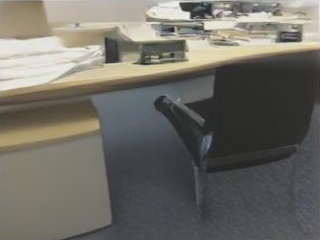}
	\captionsetup{labelformat=empty}
\end{subfigure}
\begin{subfigure}{\garbagescale\textwidth}
	\centering
	\includegraphics[width=\insidescale\textwidth]{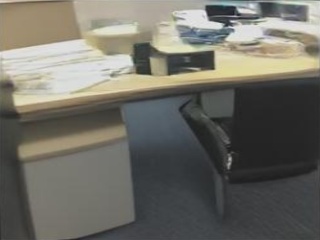}
	\captionsetup{labelformat=empty}
\end{subfigure}
\begin{subfigure}{\garbagescale\textwidth}
	\centering
	\includegraphics[width=\insidescale\textwidth]{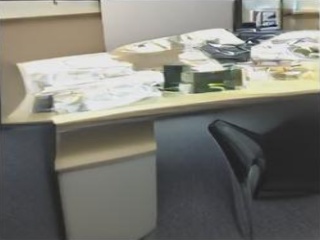}
	\captionsetup{labelformat=empty}
\end{subfigure}
\begin{subfigure}{\garbagescale\textwidth}
	\centering
	\includegraphics[width=\insidescale\textwidth]{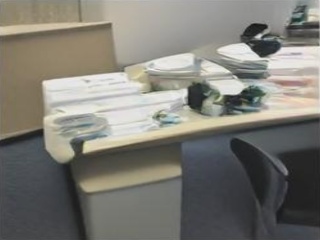}
\captionsetup{labelformat=empty}
\end{subfigure}
\begin{subfigure}{\garbagescale\textwidth}
\centering
\includegraphics[width=\insidescale\textwidth]{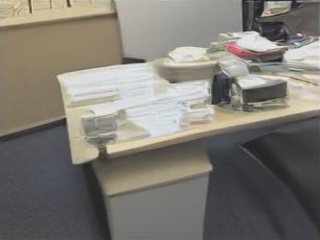}
\captionsetup{labelformat=empty}
\end{subfigure}
\begin{subfigure}{\garbagescale\textwidth}
\centering
\includegraphics[width=\insidescale\textwidth]{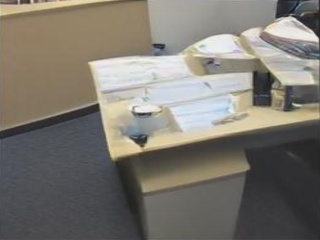}
\captionsetup{labelformat=empty}
\end{subfigure}

\vspace{5pt}

 \begin{subfigure}{\garbagescale\textwidth}
	\centering
	\includegraphics[width=\insidescale\textwidth,cfbox=red 1pt 0pt]{result/video/obs2/scene0758_00/gt/1404.jpg}
	\captionsetup{labelformat=empty}
\end{subfigure}
\begin{subfigure}{\garbagescale\textwidth}
	\centering
	\includegraphics[width=\insidescale\textwidth]{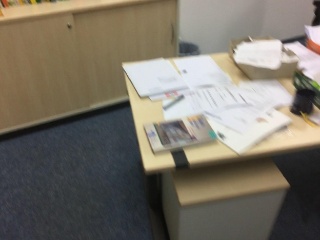}
	\captionsetup{labelformat=empty}
\end{subfigure}
\begin{subfigure}{\garbagescale\textwidth}
	\centering
	\includegraphics[width=\insidescale\textwidth]{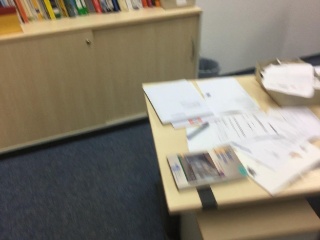}
	\captionsetup{labelformat=empty}
\end{subfigure}
\begin{subfigure}{\garbagescale\textwidth}
	\centering
	\includegraphics[width=\insidescale\textwidth]{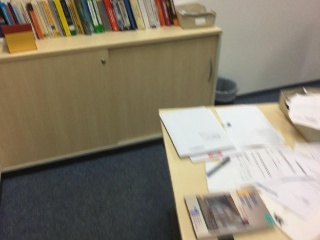}
	\captionsetup{labelformat=empty}
\end{subfigure}
\begin{subfigure}{\garbagescale\textwidth}
	\centering
	\includegraphics[width=\insidescale\textwidth]{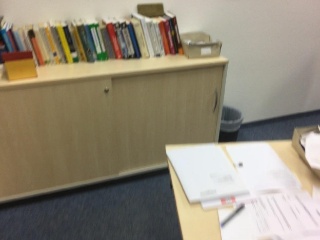}
	\captionsetup{labelformat=empty}
\end{subfigure}
\begin{subfigure}{\garbagescale\textwidth}
	\centering
	\includegraphics[width=\insidescale\textwidth]{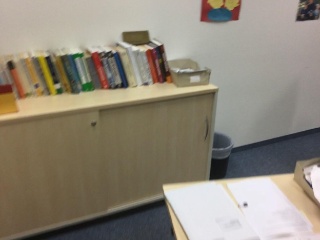}
	\captionsetup{labelformat=empty}
\end{subfigure}
\begin{subfigure}{\garbagescale\textwidth}
	\centering
	\includegraphics[width=\insidescale\textwidth]{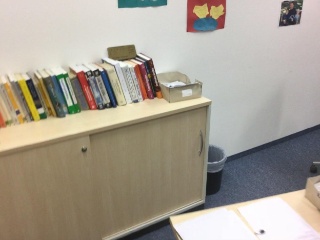}
	\captionsetup{labelformat=empty}
\end{subfigure}

\vspace{3pt}

\begin{subfigure}{\garbagescale\textwidth}
	\centering
	\includegraphics[width=\insidescale\textwidth]{result/video/obs2/scene0758_00/ours/1404.jpg}
	\captionsetup{labelformat=empty}
\end{subfigure}
\begin{subfigure}{\garbagescale\textwidth}
	\centering
	\includegraphics[width=\insidescale\textwidth]{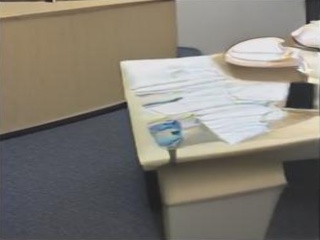}
	\captionsetup{labelformat=empty}
\end{subfigure}
\begin{subfigure}{\garbagescale\textwidth}
	\centering
	\includegraphics[width=\insidescale\textwidth]{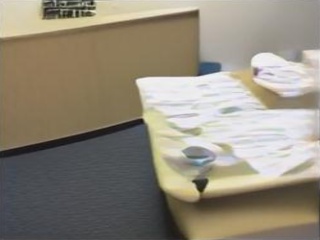}
	\captionsetup{labelformat=empty}
\end{subfigure}
\begin{subfigure}{\garbagescale\textwidth}
	\centering
	\includegraphics[width=\insidescale\textwidth]{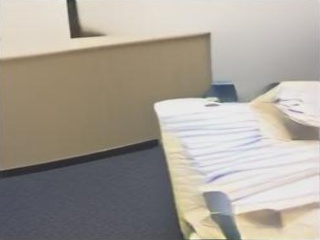}
	\captionsetup{labelformat=empty}
\end{subfigure}
\begin{subfigure}{\garbagescale\textwidth}
	\centering
	\includegraphics[width=\insidescale\textwidth]{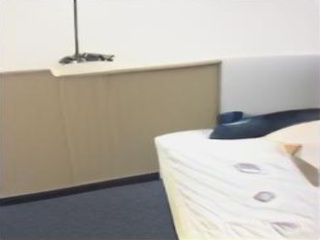}
	\captionsetup{labelformat=empty}
\end{subfigure}
\begin{subfigure}{\garbagescale\textwidth}
	\centering
	\includegraphics[width=\insidescale\textwidth]{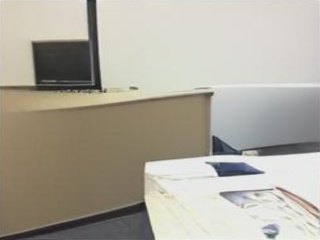}
	\captionsetup{labelformat=empty}
\end{subfigure}
\begin{subfigure}{\garbagescale\textwidth}
	\centering
	\includegraphics[width=\insidescale\textwidth]{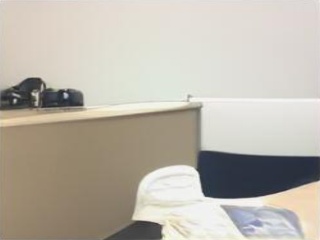}
	\captionsetup{labelformat=empty}
\end{subfigure}

 \caption{A continuous generation between only two observations (red box) and moving away. The 1st and 3rd rows are the ground truth. The 2nd and 4th rows are the generated novel views of ``in between'' and ``moving away'', respectively.} 
 \label{fig:video}
\end{figure*}

\subsection{Experimental Settings}
\noindent\textbf{Data Preparation.} We use the ScanNet dataset \cite{scannet}, following the original train/test split for training the proposed and baseline methods in the experiments. For each scan in the dataset, we randomly capture sub-scans of consecutive 256 frames. We then downsample the sub-scans to 32 frames (1/8 ratio) as a sample of a view set. For a training sample, we randomly pick 4 out of 32 frames as observed views and the rest as novel views. For a testing sample, we create 3 evaluation groups with observation number $|\mathcal{O}|$ equals 2, 4, 8, respectively. Following the settings of \cite{ddpnerf}, we hold out ``scene0708\_00'', ``scene0710\_00'', ``scene0738\_00', ``scene0758\_00'', ``scene0781\_00'' as test scenes and randomly select one sample for each scene. The comparing resolution is set to 624 × 468 after scaling and cropping dark borders. In the experiments, we assume accurate camera poses and depths so that the ground truths are provided to all the comparison methods for training and testing if necessary. We also test our trained model on the Replica dataset \cite{replica} to demonstrate the generalization ability. Due to the page limitation, we include more details in the supplementary material.

\noindent\textbf{Evaluation Metrics.} We compute the peak signal-to-noise ratio (PSNR), the structural similarity index measure (SSIM) \cite{ssim} and the learned perceptual image patch similarity (LPIPS) \cite{lpips}. We report the averaged metric results of comparing predicted novel views to ground truth images. 

\noindent\textbf{Method Setting \& Baselines.} Given a sample of $|\mathcal{O}|$ input views and 32 - $|\mathcal{O}|$ novel views, we use the $|\mathcal{O}|$ input views to build a neural point cloud for the rendered color and mask images of all 32 viewpoints. The scene context is formed by reference previews produced from the $|\mathcal{O}|$ input views and probed previews from the 32-$|\mathcal{O}|$-1 novel views. The query is produced from the rest 1 novel view. The neural geometry module follows the Point-NeRF settings \cite{pointnerf}. The image converter module follows the settings in \cite{lookout}. The network of the view generator module includes an encoder and a decoder. The encoder architecture mainly follows \cite{srt} with slight modifications. The decoder is a stack of 6 vanilla transformer decoder layers. We train the model with a learning rate 1e-4 and batch size 16 using the Adam optimizer. We set up five baseline methods for comparison: Point-NeRF \cite{pointnerf}, PixelSynth \cite{pixelsynth}, IBRNet \cite{ibrnet}, ViewFormer \cite{viewformer}, and NeRF with dense depth priors (DDP-NeRF) \cite{ddpnerf}. For Point-NeRF, we train its MVSNet and ray marching MLPs across scenes, and test it in a feed-forward manner. For PixelSynth, we take the pre-trained model provided by the authors, and predict novel views by choosing the nearest observation as the start image to outpaint its re-projection of the novel pose. For IBRNet and ViewFormer, we train and test models with the same setting to our method (IBRNet is tested in feed-forward). For DDP-NeRF, we optimize the model with ground truth depths and camera poses for each scene.

\subsection{Primary Results \& Analyses}
We compare the quantitative results on 3 groups of sparse input views with observation number $|\mathcal{O}|$ = 2, 4, 8, respectively. As the results presented in Table \ref{tab:primary}, our method outperforms all the baselines on PSNR, SSIM, and LPIPS. Note that, without ``Geometry'' (i.e., rendered color images and masks from the neural geometry module), the performance of our method significantly drops. This proves the importance of 3D structure for generating high-quality novel views. We show the generations of novel views with ground truths in Figure \ref{fig:obs2}, \ref{fig:obs4}, and \ref{fig:obs8} (more results in supplementary). The results of Point-NeRF are often corrupted and scattered, which are caused by the incomplete underlying point clouds. Without generalization ability, Point-NeRF can hardly perform well with sparse inputs. PixelSynth produces distorted views as the poses of the novel views are largely shifted from the referenced observations. Therefore, the reasoning 3D surfaces cannot be projected correctly which causes the distortion. The results of IBRNet are often blurred and show black areas where rays hit no clue from sparse observations. ViewFormer generates vague shapes but lacks details as it only depends on compressed code where information is lost. DDP-NeRF performs the best among all the baselines. But due to the large sparsity, the renderings of DDP-NeRF unavoidably overfit to input views that cause blurs even with depth priors. Our method generally outperforms the others in terms of fidelity and details. With increasing the observation number, our method generates novel views of better visual quality overall (proved by metrics), sufficiently leveraging the given information and demonstrating strong applicability.

\subsection{View Consistency}
To demonstrate the view consistency of SparseGNV, we show continuous novel view generations between two observations in Figure \ref{fig:video} (first two rows). The quality and consistency are fairly maintained without significant perturbation. The neural geometry module provides a strong scene context of 3D structure, which ensures a stable generation ability by the downstream modules. We further show a sequence of generated novel views that moves away from the two observations in Figure \ref{fig:video} (last two rows). The office desk is fairly maintained until moving out, as there is enough clue of its shape and appearance. Unfortunately, the cabinet appears with only its surface, and the books on top of the cabinet are completely missed of generation. Since there is no clue of their occurrence, the model tends to generate a white wall to maintain consistency.

\subsection{Time \& Memory \& Model Size}
We conduct experiments on NVIDIA V100 GPUs. The inference speed of the trained model is 0.83s per batch of 24 images. The training of the neural geometry module takes about 1 day using 1 GPU (batch size 1, memory $\leq$ 20G depends on scene size). The training of the image generator module takes about 1 week using 2 GPUs (batch size 16, 20.9G). VQ decoder uses the pre-trained checkpoint from~\cite{lookout} thus no re-training is required. The total parameter count of the three modules is: 0.724M (Point-NeRF) + 88M (convolution and transformer network) + 76M (VQ decoder). Please note that the training needs to be performed only once, and the trained framework can generalize to unseen indoor scenes without further fine-tuning.

\section{Conclusions \& Limitations}
In this paper, we study the problem of novel view synthesis of indoor scenes given sparse input views. To generate both photorealistic and consistent novel views, we propose SparseGNV: a learning framework that incorporates 3D structure into image generative models. The framework is designed with three network-based modules. The neural geometry module builds a 3D neural point cloud to produce rendered images from arbitrary viewpoints. The view generator module takes the rendered images to form scene context and query, which are fed into a convolution and transformer-based network to generate the target novel view represented in VQ codebook tokens. The image converter module finally reconstructs the tokens back to the view image. SparseGNV is trained across scenes to learn priors, and infers novel views of unseen scenes in a feed-forward manner. The evaluation results on real-world and synthetic indoor scenes demonstrate the exceeding performance of the method over recent baselines.
\\~
\textbf{Limitations.} SparseGNV synthesizes novel views using an image generation model based on the VQ codebook. The output is therefore less stable compared to the volume rendering-based methods. For example, the object details and lighting could be altered. The framework also requires camera poses and depths which can be unavailable when the observed views are extremely sparse.

\pagebreak

{\small
\bibliographystyle{ieee_fullname}
\bibliography{arxiv}

\begin{thebibliography}{10}\itemsep=-1pt

\bibitem{mipnerf}
Jonathan~T. Barron, Ben Mildenhall, Matthew Tancik, Peter Hedman, Ricardo
  Martin-Brualla, and Pratul~P. Srinivasan.
\newblock Mip-nerf: A multiscale representation for anti-aliasing neural
  radiance fields.
\newblock {\em Computer Vision and Pattern Recognition}, 2021.

\bibitem{mvsnerf}
Anpei Chen, Zexiang Xu, Fuqiang Zhao, Xiaoshuai Zhang, Fanbo Xiang, Jingyi Yu,
  and Hao Su.
\newblock Mvsnerf: Fast generalizable radiance field reconstruction from
  multi-view stereo.
\newblock {\em International Conference on Computer Vision}, 2021.

\bibitem{geoaug}
Di Chen, Yu Liu, Lianghua Huang, Bin Wang, and Pan Pan.
\newblock Geoaug: Data augmentation for few-shot nerf with geometry
  constraints.
\newblock {\em European Conference on Computer Vision}, pages 322--337, 2022.

\bibitem{scannet}
Angela Dai, Angel~X Chang, Manolis Savva, Maciej Halber, Thomas Funkhouser, and
  Matthias Nie{\ss}ner.
\newblock Scannet: Richly-annotated 3d reconstructions of indoor scenes.
\newblock {\em Computer Vision and Pattern Recognition}, pages 5828--5839,
  2017.

\bibitem{spsg}
Angela Dai, Yawar Siddiqui, Justus Thies, Julien Valentin, and Matthias
  Nie{\ss}ner.
\newblock Spsg: Self-supervised photometric scene generation from rgb-d scans.
\newblock {\em Computer Vision and Pattern Recognition}, 2021.

\bibitem{bert}
Jacob Devlin, Ming-Wei Chang, Kenton Lee, and Kristina Toutanova.
\newblock Bert: Pre-training of deep bidirectional transformers for language
  understanding.
\newblock {\em Conference of the North American Chapter of the Association for
  Computational Linguistics: Human Language Technologies}, 2019.

\bibitem{synsin}
Georgia Gkioxari, Olivia Wiles, Richard Szeliski, Justin Johnson, Olivia Wiles,
  Georgia Gkioxari, Richard Szeliski, and Justin Johnson.
\newblock Synsin: End-to-end view synthesis from a single image.
\newblock {\em Computer Vision and Pattern Recognition}, 2019.

\bibitem{StevenJGortler1996TheL}
Steven~J. Gortler, Radek Grzeszczuk, Richard Szeliski, and Michael Cohen.
\newblock The lumigraph.
\newblock {\em International Conference on Computer Graphics and Interactive
  Techniques}, 1996.

\bibitem{hedman2016scalable}
Peter Hedman, Tobias Ritschel, George Drettakis, and Gabriel Brostow.
\newblock Scalable inside-out image-based rendering.
\newblock {\em ACM Transactions on Graphics}, 35(6):1--11, 2016.

\bibitem{dreamfields}
Anubhav Jain, Ben Mildenhall, Jonathan~T. Barron, Pieter Abbeel, and Ben Poole.
\newblock Zero-shot text-guided object generation with dream fields.
\newblock {\em Computer Vision and Pattern Recognition}, 2021.

\bibitem{dietnerf}
Ajay Jain, Matthew Tancik, and Pieter Abbeel.
\newblock Putting nerf on a diet: Semantically consistent few-shot view
  synthesis.
\newblock {\em Computer Vision and Pattern Recognition}, 2021.

\bibitem{infonerf}
Mijeong Kim, Seonguk Seo, and Bohyung Han.
\newblock Infonerf: Ray entropy minimization for few-shot neural volume
  rendering.
\newblock {\em Computer Vision and Pattern Recognition}, pages 12912--12921,
  2022.

\bibitem{pathdreamer}
Jing~Yu Koh, Honglak Lee, Yinfei Yang, Jason Baldridge, and Peter Anderson.
\newblock Pathdreamer: A world model for indoor navigation.
\newblock {\em International Conference on Computer Vision}, 2021.

\bibitem{viewformer}
Jon{\'a}{\vs} Kulh{\'a}nek, Erik Derner, Torsten Sattler, and Robert
  Babu{\vs}ka.
\newblock Viewformer: Nerf-free neural rendering from few images using
  transformers.
\newblock {\em European Conference on Computer Vision}, 2022.

\bibitem{gss}
Jiabao Lei, Jiapeng Tang, and Kui Jia.
\newblock Generative scene synthesis via incremental view inpainting using rgbd
  diffusion models, 2022.

\bibitem{MarcLevoy1996LightFR}
Marc Levoy and Pat Hanrahan.
\newblock Light field rendering.
\newblock {\em International Conference on Computer Graphics and Interactive
  Techniques}, 1996.

\bibitem{compnvs}
Zuoyue Li, Tianxing Fan, Zhenqiang Li, Zhaopeng Cui, Yoichi Sato, Marc
  Pollefeys, and Martin~R. Oswald.
\newblock Compnvs: Novel view synthesis with scene completion.
\newblock 2022.

\bibitem{nsff}
Zhengqi Li, Simon Niklaus, Noah Snavely, and Oliver Wang.
\newblock Neural scene flow fields for space-time view synthesis of dynamic
  scenes.
\newblock {\em Computer Vision and Pattern Recognition}, 2020.

\bibitem{barf}
Chen-Hsuan Lin, Wei-Chiu Ma, Antonio Torralba, and Simon Lucey.
\newblock Barf: Bundle-adjusting neural radiance fields.
\newblock {\em International Conference on Computer Vision}, 2021.

\bibitem{nerfw}
Ricardo Martin-Brualla, Noha Radwan, Mehdi S.~M. Sajjadi, Jonathan~T. Barron,
  Alexey Dosovitskiy, and Daniel Duckworth.
\newblock Nerf in the wild: Neural radiance fields for unconstrained photo
  collections.
\newblock {\em Computer Vision and Pattern Recognition}, 2020.

\bibitem{gnerf}
Quan Meng, Anpei Chen, Haimin Luo, Minye Wu, Hao Su, Lan Xu, Xuming He, and
  Jingyi Yu.
\newblock Gnerf: Gan-based neural radiance field without posed camera.
\newblock {\em Computer Vision and Pattern Recognition}, 2021.

\bibitem{nerf}
Ben Mildenhall, Pratul~P. Srinivasan, Matthew Tancik, Jonathan~T. Barron, Ravi
  Ramamoorthi, and Ren Ng.
\newblock Nerf: Representing scenes as neural radiance fields for view
  synthesis.
\newblock {\em European Conference on Computer Vision}, 2020.

\bibitem{regnerf}
Michael Niemeyer, Jonathan~T. Barron, Ben Mildenhall, Mehdi S.~M. Sajjadi,
  Andreas Geiger, and Noha Radwan.
\newblock Regnerf: Regularizing neural radiance fields for view synthesis from
  sparse inputs.
\newblock {\em Computer Vision and Pattern Recognition}, 2021.

\bibitem{vqgan}
Bj{\"o}rn Ommer, Patrick Esser, Robin Rombach, Patrick Esser, Robin Rombach,
  and Bj{\"o}rn Ommer.
\newblock Taming transformers for high-resolution image synthesis.
\newblock {\em Computer Vision and Pattern Recognition}, 2020.

\bibitem{spade}
Taesung Park, Ming-Yu Liu, Ting-Chun Wang, and Jun-Yan Zhu.
\newblock Semantic image synthesis with spatially-adaptive normalization.
\newblock {\em Computer Vision and Pattern Recognition}, 2019.

\bibitem{philip2021free}
Julien Philip, S{\'e}bastien Morgenthaler, Micha{\"e}l Gharbi, and George
  Drettakis.
\newblock Free-viewpoint indoor neural relighting from multi-view stereo.
\newblock {\em ACM Transactions on Graphics (TOG)}, 40(5):1--18, 2021.

\bibitem{dreamfusion}
Ben Poole, Ajay Jain, Jonathan~T. Barron, and Ben Mildenhall.
\newblock Dreamfusion: Text-to-3d using 2d diffusion.
\newblock {\em arXiv}, 2022.

\bibitem{dnerf}
Albert Pumarola, Enric Corona, Gerard Pons-Moll, and Francesc Moreno-Noguer.
\newblock D-nerf: Neural radiance fields for dynamic scenes.
\newblock {\em Computer Vision and Pattern Recognition}, 2020.

\bibitem{dalle}
Aditya Ramesh, Mikhail Pavlov, Gabriel Goh, Scott Gray, Chelsea Voss, Alec
  Radford, Mark Chen, and Ilya Sutskever.
\newblock Zero-shot text-to-image generation.
\newblock {\em International Conference on Machine Learning}, 2021.

\bibitem{lookout}
Xuanchi Ren, Hkust Xiaolong, Wang Uc, and San Diego.
\newblock Look outside the room: Synthesizing a consistent long-term 3d scene
  video from a single image.
\newblock {\em Computer Vision and Pattern Recognition}, 2022.

\bibitem{pixelsynth}
Chris Rockwell, David~F. Fouhey, and Justin Johnson.
\newblock Pixelsynth: Generating a 3d-consistent experience from a single
  image.
\newblock {\em International Conference on Computer Vision}, 2021.

\bibitem{ddpnerf}
Barbara Roessle, Jonathan~T. Barron, Ben Mildenhall, Pratul~P. Srinivasan, and
  Matthias Niessner.
\newblock Dense depth priors for neural radiance fields from sparse input
  views.
\newblock {\em Computer Vision and Pattern Recognition}, 2022.

\bibitem{srt}
Mehdi S.~M. Sajjadi, Henning Meyer, Etienne Pot, Urs Bergmann, Klaus Greff,
  Noha Radwan, Suhani Vora, Mario Lucic, Daniel Duckworth, Alexey Dosovitskiy,
  Jakob Uszkoreit, Tom Funkhouser, and Andrea Tagliasacchi.
\newblock Scene representation transformer: Geometry-free novel view synthesis
  through set-latent scene representations.
\newblock {\em Computer Vision and Pattern Recognition}, 2022.

\bibitem{srn}
Vincent Sitzmann, Michael Zollh{\"o}fer, and Gordon Wetzstein.
\newblock Scene representation networks: Continuous 3d-structure-aware neural
  scene representations.
\newblock {\em Advances in Neural Information Processing Systems}, 32, 2019.

\bibitem{replica}
Julian Straub, Thomas Whelan, Lingni Ma, Yufan Chen, Erik Wijmans, Simon Green,
  Jakob~J. Engel, Raul Mur-Artal, Carl Ren, Shobhit Verma, Anton Clarkson,
  Mingfei Yan, Brian Budge, Yajie Yan, Xiaqing Pan, June Yon, Yuyang Zou,
  Kimberly Leon, Nigel Carter, Jesus Briales, Tyler Gillingham, Elias Mueggler,
  Luis Pesqueira, Manolis Savva, Dhruv Batra, Hauke~M. Strasdat, Renzo~De
  Nardi, Michael Goesele, Steven Lovegrove, and Richard Newcombe.
\newblock The {R}eplica dataset: A digital replica of indoor spaces.
\newblock {\em arXiv}, 2019.

\bibitem{vqvae}
Aaron van~den Oord, Oriol Vinyals, and Koray Kavukcuoglu.
\newblock Neural discrete representation learning.
\newblock {\em Advances in Neural Information Processing Systems}, 2017.

\bibitem{ibrnet}
Qianqian Wang, Zhicheng Wang, Kyle Genova, Pratul~P. Srinivasan, Howard Zhou,
  Jonathan~T. Barron, Ricardo Martin-Brualla, Noah Snavely, and Thomas
  Funkhouser.
\newblock Ibrnet: Learning multi-view image-based rendering.
\newblock {\em Computer Vision and Pattern Recognition}, 2021.

\bibitem{ssim}
Zhou Wang, Alan~C Bovik, Hamid~R Sheikh, and Eero~P Simoncelli.
\newblock Image quality assessment: from error visibility to structural
  similarity.
\newblock {\em IEEE Transactions on Image Processing}, 13(4):600--612, 2004.

\bibitem{pointnerf}
Qiangeng Xu, Zexiang Xu, Julien Philip, Sai Bi, Zhixin Shu, Kalyan Sunkavalli,
  and Ulrich Neumann.
\newblock Point-nerf: Point-based neural radiance fields.
\newblock {\em Computer Vision and Pattern Recognition}, pages 5438--5448,
  2022.

\bibitem{mvsnet}
Yao Yao, Zixin Luo, Shiwei Li, Tian Fang, and Long Quan.
\newblock Mvsnet: Depth inference for unstructured multi-view stereo.
\newblock {\em European Conference on Computer Vision}, 2018.

\bibitem{plenoctrees}
Alex Yu, Ruilong Li, Matthew Tancik, Hao Li, Ren Ng, and Angjoo Kanazawa.
\newblock Plenoctrees for real-time rendering of neural radiance fields.
\newblock {\em International Conference on Computer Vision}, 2021.

\bibitem{pixelnerf}
Alex Yu, Vickie Ye, Matthew Tancik, and Angjoo Kanazawa.
\newblock pixelnerf: Neural radiance fields from one or few images.
\newblock {\em Computer Vision and Pattern Recognition}, 2021.

\bibitem{lpips}
Richard Zhang, Phillip Isola, Alexei~A Efros, Eli Shechtman, and Oliver Wang.
\newblock The unreasonable effectiveness of deep features as a perceptual
  metric.
\newblock {\em Computer Vision and Pattern Recognition}, pages 586--595, 2018.

\end{thebibliography}
}

\end{document}